\documentclass[11pt, dvipsnames, svgnames, x11names]{article}

\usepackage[preprint]{acl}

\usepackage{times}
\usepackage{latexsym}

\usepackage[T1]{fontenc}

\usepackage[utf8]{inputenc}

\usepackage{microtype}

\usepackage{inconsolata}

\usepackage{graphicx}

\usepackage{enumitem}
\usepackage{amsmath}
\usepackage{algorithm}
\usepackage{algorithmicx}
\usepackage{algpseudocode}
\usepackage{multirow}
\usepackage{booktabs}
\usepackage{pifont}
\newcommand{\cmark}{\ding{51}} 
\newcommand{\xmark}{\ding{55}} 
\usepackage{makecell}
\usepackage[normalem]{ulem}
\useunder{\uline}{\ul}{}

\usepackage{tcolorbox}
\usepackage{colortbl}
\tcbuselibrary{breakable}
\usepackage{wrapfig}

\definecolor{bg_sft}{RGB}{255, 245, 245} 
\definecolor{bg_rl}{RGB}{245, 255, 245}  
\definecolor{border_gray}{RGB}{200, 200, 200}

\title{Writing-RL: Advancing Long-form Writing via \\ Adaptive Curriculum Reinforcement Learning}

\author{
 \textbf{Xuanyu Lei\textsuperscript{1,2}},
 \textbf{Chenliang Li\textsuperscript{3}},
 \textbf{Yuning Wu\textsuperscript{3}},
 \textbf{Kaiming Liu\textsuperscript{2}}, 
 \textbf{Weizhou Shen\textsuperscript{3}}, \\
 \textbf{Peng Li\textsuperscript{1,$\dagger$}},
 \textbf{Ming Yan\textsuperscript{3,$\dagger$}},
 \textbf{Fei Huang\textsuperscript{3}}, 
 \textbf{Ya-Qin Zhang\textsuperscript{1}},
 \textbf{Yang Liu\textsuperscript{1,2,$\dagger$}} \\
 \textsuperscript{1}Institute for AI Industry Research (AIR), Tsinghua University, Beijing, China\\
 \textsuperscript{2}Dept. of Comp. Sci. \& Tech., Institute for AI, Tsinghua University, Beijing, China\\
 \textsuperscript{3}Institute of Intelligent Computing, Alibaba Group\\
 \texttt{leixy24@mails.tsinghua.edu.cn, lipeng@air.tsinghua.edu.cn} \\
 \texttt{ym119608@alibaba-inc.com, liuyang2011@tsinghua.edu.cn}
}

\begin{document}
\maketitle

\begin{abstract}

Recent advances in Large Language Models (LLMs) have enabled strong performance in long-form writing, but current training paradigms remain limited: Supervised Fine-Tuning (SFT) remains constrained by data saturation and performance ceilings, while Reinforcement Learning with Verifiable Reward (RLVR), though successful in verifiable domains like math and code, cannot be directly migrated to open-ended long-form writing due to a lack of ground-truths.
To further advance long-form writing, we present \textbf{Writing-RL}: an \textit{Adaptive Curriculum Reinforcement Learning} framework to advance long-form writing capabilities beyond SFT. 
The framework consists of three key components: \textit{Margin-aware Data Selection} strategy that prioritizes samples with high learning potential, \textit{Pairwise Comparison Reward} mechanism that provides discriminative learning signals in the absence of verifiable rewards, and \textit{Dynamic Reference Scheduling} approach, which plays a critical role by adaptively adjusting task difficulty based on evolving model performance.
Experiments on 7B-scale writer models show that Writing-RL effectively improves long-form writing performance over strong SFT baselines.
Furthermore, we observe that models trained with long-output RL generalize surprisingly well to long-input reasoning tasks, potentially offering a promising perspective for rethinking long-context training.

\end{abstract}

\noindent\let\thefootnote\relax\footnotetext{$^\dagger$ Corresponding Authors.}
\noindent\let\thefootnote\relax\footnotetext{$^\ddagger$ Code is released at \url{https://github.com/Tongyi-Zhiwen/Writing-RL}.}

\section{Introduction}
Recent years have witnessed the remarkable advance of Large Language Models (LLMs)~\citep{DBLP:journals/corr/abs-2303-08774, DBLP:journals/corr/abs-2501-12948, DBLP:journals/corr/abs-2303-18223} to follow complicated instructions and provide helpful responses.
Among their impressive capabilities, long-form writing, which aims to generate long and high-quality articles, has drawn increasing attention~\citep{DBLP:journals/corr/abs-2503-04723, DBLP:journals/corr/abs-2408-07055, DBLP:journals/corr/abs-2503-05244} due to its broad practical applications.

However, generating articles of both sufficient length and high quality is non-trivial for current LLMs. 
Previous research has identified several challenges in employing LLMs for long-form generation, including inherently limited output ceiling~\citep{DBLP:journals/corr/abs-2408-07055, DBLP:journals/corr/abs-2502-14834} and performance degradation as output length grows~\citep{DBLP:journals/corr/abs-2503-05244, DBLP:journals/corr/abs-2502-14834}.
To address these issues, recent efforts perform targeted Supervised Fine-Tuning (SFT) on LLMs to extend their output lengths, with long-generation datasets constructed by iterative agent pipelines~\citep{DBLP:journals/corr/abs-2408-07055, DBLP:journals/corr/abs-2410-23933, DBLP:journals/corr/abs-2503-05244} or instruction back-translation~\citep{DBLP:conf/emnlp/PhamSI24, DBLP:journals/corr/abs-2401-17268}.
Though effective, these approaches introduce a heavy burden of dataset construction due to the broad coverage of writing tasks and potential copyright issues~\citep{DBLP:conf/acl/MainiSBG0J24} when incorporating human-written texts. Furthermore, training LLMs to imitate the collected long-generation responses inherently imposes a capability upper bound determined by teacher models or human experts, which may cause data saturation and sample inefficiency.

\begin{figure*}[t]
    \centering
    \includegraphics[width=\linewidth]{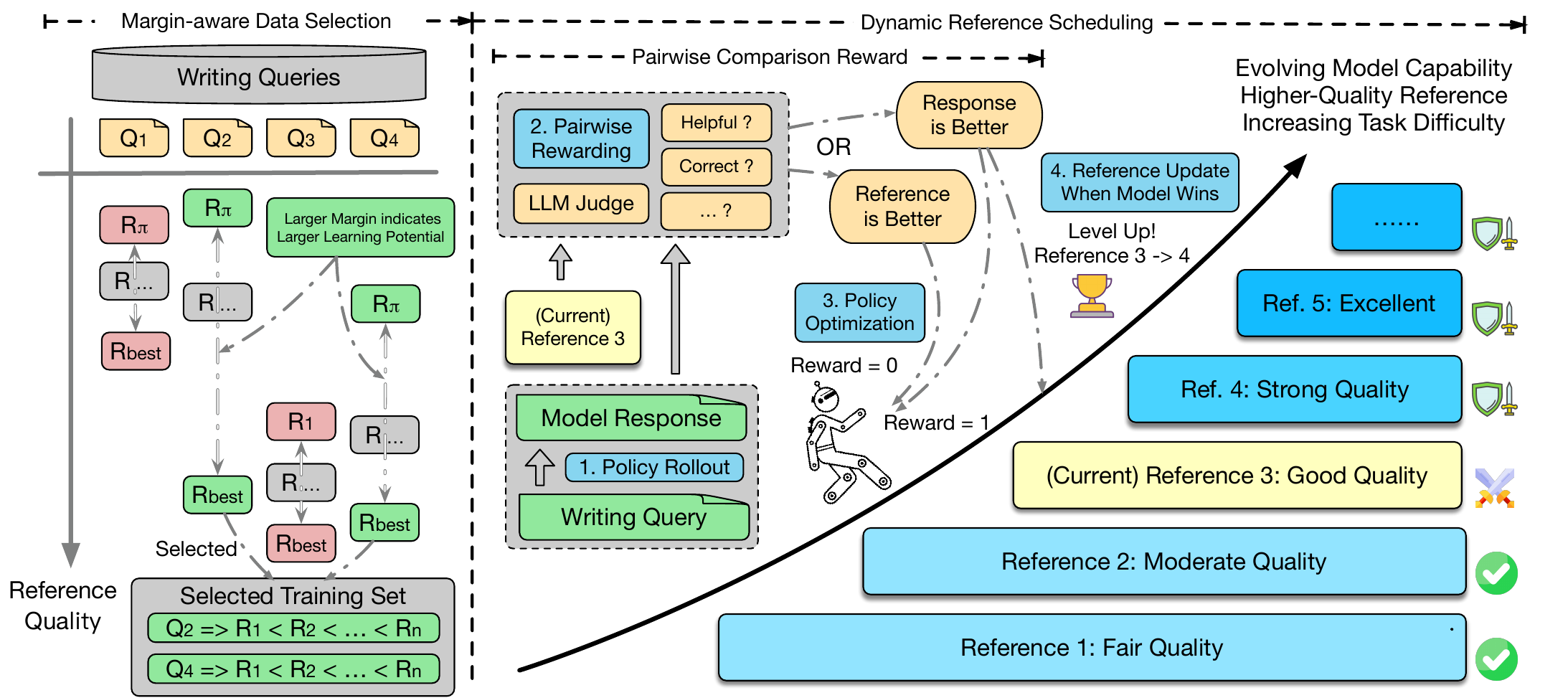}
    \vspace{-2mm}
    \caption{Overall framework of \textbf{Writing-RL}. 1) \textit{Margin-aware Data Selection}: prioritizes samples with high learning potential; 2) \textit{Pairwise Comparison Reward}: provides more discriminative reward signals; 3) \textit{Dynamic Reference Scheduling}: adaptively incentivizes the model to surpass progressively stronger references.}
    \label{fig:framework}
    \vspace{-5mm}
\end{figure*}

Meanwhile, recent progress of Reinforcement Learning (RL) with Verifiable Rewards~\citep{DBLP:journals/corr/abs-2501-12948, DBLP:journals/corr/abs-2501-12599, yuan2025vapo} in reasoning-intensive areas reveals a promising direction to advance model capabilities beyond SFT. 
In long-form writing, however, the lack of ground truths prevents a straightforward transfer of these successes. \citet{DBLP:journals/corr/abs-2506-18841} utilize static reward models for grading, failing to dynamically adapt to evolving model capability.
Overall, adaptive online RL for long-form writing remains under-explored and presents several challenges:

\begin{itemize}[leftmargin=1.5em,itemsep=0pt,parsep=0.2em,topsep=0.1em,partopsep=0.0em]
\item \textbf{Data Selection}: Data quality and difficulty play a critical role in eliciting model potential. However, the optimal approach for selecting data for RL in long-form writing tasks remains unclear, requiring more explorations towards better learning efficiency.
\item \textbf{Reward Design}: Rule-based outcome rewards~\citep{DBLP:journals/corr/abs-2501-12948} cannot be directly applied to generative writing tasks. Without ground-truth labels, constructing an effective reward mechanism for long-form writing poses a significant challenge.
\item \textbf{Curriculum Scheduling}: Curriculum Learning~\citep{DBLP:conf/icml/BengioLCW09} is widely used to progressively improve model performance, but current static scheduling fails to adapt to the model's evolving competence, thereby reducing training effectiveness.
\end{itemize}

To tackle these challenges, our work proposes \textbf{Writing-RL}: an \textit{Adaptive Curriculum Reinforcement Learning} framework tailored for long-form writing.
As illustrated in Figure~\ref{fig:framework}, our framework begins with \textit{\textbf{Margin-aware Data Selection}} strategy which leverages the quality differential between the policy model response and the highest-quality reference as a measure of \textit{learning potential}, diverging from the conventional difficulty-prioritized selection approach. 
Considering the limited discriminative capacity of pointwise rewarding, we construct a \textit{\textbf{Pairwise Comparison Reward}} mechanism which challenges the policy model to generate responses of better quality than provided references to earn positive rewards. 
To facilitate progressive model enhancement, we propose a \textbf{\textit{Dynamic Reference Scheduling}} approach that assigns each query a set of references with progressively increasing quality. 
The scheduling approach dynamically updates the references per sample when the evolving policy model surpasses the current reference during training. 
In this way, the dynamic curriculum adjusts sample-level task difficulty based on the current model performance, encouraging the model to consistently outperform a marginally superior reference. 
This rationale aligns with insights from recent R1-like RL practices~\citep{shi2025efficient, bae2025online} that samples neither too easy nor too difficult help to achieve the best learning efficiency. 

To evaluate our framework, we conduct continuous reinforcement training on top of supervised fine-tuned writer models. 
The results indicate that our RL framework effectively boosts the long-form writing capability, advancing the SOTA performances of 7B-level writer models. 
Besides the improvement in long-form generation, we also observe an inspiring generalization phenomenon: our RL-trained writer model \textit{(average input length < 1k)} shows a surprising improvement in long-text reasoning tasks \textit{(input length: 8k–2M)}, in contrast to the performance degradation of the SFT-trained model. 
The results suggest a novel perspective on long-context learning: models trained on \textit{long-output} tasks may also improve their reasoning abilities on \textit{long-input} tasks, offering new insights into the relationship between long-context understanding and generation.

In summary, the contributions of our work are:
\begin{itemize}[leftmargin=1.5em,itemsep=0pt,parsep=0.2em,topsep=0.1em,partopsep=0.0em]
\item We propose \textbf{Writing-RL}: an \textit{Adaptive Curriculum Reinforcement Learning} framework for long-form writing, which integrates three key components: \textit{Margin-aware Data Selection}, \textit{Pairwise Comparison Reward}, and \textit{Dynamic Reference Scheduling}.
\item Particularly, we propose \textbf{Dynamic Reference Scheduling}, which adaptively adjusts sample-level task difficulty based on the model’s evolving performance. This dynamic curriculum encourages the model to continually outperform progressively stronger references.
\item Our writer model achieves state-of-the-art performance at its scale, demonstrating the effectiveness of Writing-RL. Furthermore, we observe inspiring \textbf{Output-to-Input Generalization} from \textit{long-output} generation to \textit{long-input} reasoning, potentially offering a novel perspective to rethink long-context training.

\end{itemize}

\section{Related Work}

\noindent\textbf{Training Methods for Long-form Writing.}
Recent efforts to advance long-form writing capabilities~\citep{DBLP:journals/corr/abs-2408-07055, DBLP:journals/corr/abs-2503-05244} mainly focus on constructing long-generation datasets for fine-tuning. Main approaches include teacher model distillation~\citep{DBLP:journals/corr/abs-2503-05244}, iterative agent pipelines for extended output~\citep{DBLP:journals/corr/abs-2408-07055, DBLP:journals/corr/abs-2502-14834, DBLP:journals/corr/abs-2410-23933} and instruction back-translation~\citep{DBLP:conf/emnlp/PhamSI24, DBLP:journals/corr/abs-2401-17268}. \cite{DBLP:journals/corr/abs-2506-18841} incorporates static reward models for supervision, which fails to dynamically adapt to evolving model capability during training.
However, the application of adaptive online reinforcement learning methods are relatively underexplored, hindering further improvement.

\noindent\textbf{Long-form Writing Evaluation.}
Long-form writing~\citep{DBLP:journals/corr/abs-2503-04723} requires LLMs to write open-ended articles, posing challenges for evaluation due to the lack of ground-truths. 
Earlier studies establish writing benchmarks~\citep{DBLP:journals/corr/abs-2503-05244, DBLP:journals/corr/abs-2409-16191}, with proprietary models~\citep{DBLP:journals/corr/abs-2408-07055, DBLP:journals/corr/abs-2312-06281, DBLP:conf/acl/LiuLWHFWCKXTZ0G24} or fine-tuned LLMs~\citep{DBLP:journals/corr/abs-2503-05244, DBLP:conf/acl/KeWFLLCWZDWTH24} to serve as judges. 
However, there exist several biases of including position bias and self-enhancement bias~\citep{DBLP:conf/nips/ZhengC00WZL0LXZ23}, challenging the reliability of LLM-as-Judge evaluation methods.

\noindent\textbf{Curriculum Learning.} Reinforcement Learning methods~\citep{DBLP:journals/corr/SchulmanWDRK17, DBLP:journals/corr/abs-2402-03300, DBLP:journals/corr/abs-2501-12948} have become a critical step to elicit LLM capabilities. 
To boost efficiency, Curriculum Learning~\citep{DBLP:conf/icml/BengioLCW09} has been widely adopted in RL practices~\citep{DBLP:journals/corr/abs-2501-12599, DBLP:journals/corr/abs-2502-14768, DBLP:journals/corr/abs-2503-10460}, including static difficulty-based scheduling~\citep{deepscaler2025, DBLP:journals/corr/abs-2503-17287} and dynamic data selection~\citep{bae2025online, shi2025efficient}. However, they use rule-based correctness as a measure for difficulty and perform sample selection, which increases rollouts and may cause imbalanced learning across samples.

\section{Writing-RL}
\label{sec:method}

In this work, we propose \textbf{Writing-RL}, an \textit{Adaptive Curriculum Reinforcement Learning} framework aimed at further improving long-form writing capabilities. 
The framework comprises three key components: \textit{Margin-aware Data Selection}, \textit{Pairwise Comparison Reward} and \textit{Dynamic Reference Scheduling}. 
We describe these components below.

\subsection{Margin-aware Data Selection}
\label{sec:data_selection}

Previous data selection approaches typically take question difficulty as a key criterion, measured by the accuracy of the policy model~\citep{shi2025efficient, bae2025online}, simplistic indicators~\citep{DBLP:conf/acl/ChengLLZLLZ20, DBLP:journals/corr/abs-2503-10105} like solution step counts or simple heuristics grounded in human intuition~\citep{DBLP:conf/nips/HendrycksBKABTS21}.
While difficulty-prioritized data selection has been effective in verifiable tasks such as math and code, it depends on clearly defined ground truth to measure difficulty. In open-ended writing tasks, however, the lack of ground-truths makes difficulty hard to measure.

To address this issue, we propose \textit{Margin-aware Data Selection}, which uses the performance gap between the policy output and the highest-quality reference as a measure of \textit{learning potential}. Our intuition is simple: a question suitable for learning is a question with sufficient room for performance improvement. Specifically, the procedure is detailed as follows.

\noindent\textbf{Generation with Multiple LLMs. } 
Instead of relying on a single model as the difficulty estimator~\citep{shi2025efficient, bae2025online}, we leverage a set of competitive LLMs \(\mathcal{C} = \{\pi, M_1, M_2, \dots\}\), including the policy model $\pi$, to generate diverse candidate responses for each writing instruction.

\noindent\textbf{Multi-dimensional Grading. } 
Each generated response \( r_j \) from model \( M_j \in \mathcal{C} \) is graded using a multi-dimensional pointwise LLM-as-a-Judge approach~\citep{DBLP:conf/acl/LiuLWHFWCKXTZ0G24, DBLP:journals/corr/abs-2503-05244}, with averaged quality score denoted as \( s_j \) per response.

\noindent\textbf{Data Selection on Learning Potential. } 
To prioritize samples from which the policy model can benefit most, we define the \textit{model-grounded learning potential} \( p \) as the quality gap between the best competitor and the policy model:

\vspace{-2mm}

{\small
\[
p = \max_{j \in \mathcal{C},\; j \ne \pi} (s_j - s_\pi)
\]
}

\vspace{-2mm}

\noindent where \( s_\pi \) is the score of the policy model's response. A higher \( p \) indicates greater headroom for improvement. To filter out potential noise, we first discard samples where all the competitors receive low scores, as such instructions are often overly difficult or suffer from quality issues themselves. After filtering, we retain the top-\(k\) examples based on their learning potential \( p \) to construct the training set.

\subsection{Pairwise Comparison Reward Mechanism}
\label{sec:pairwise_reward}

The reward function is a critical component to guide policy optimization in RL. While rule-based outcome reward~\citep{DBLP:journals/corr/abs-2501-12948, DBLP:journals/corr/abs-2501-12599} has been proven to be remarkably effective in verifiable tasks, it cannot be directly applied to long-form writing due to the lack of ground truth and its subjective nature, posing challenges to reward design.

Recent efforts utilize LLM-as-a-Judge~\citep{DBLP:conf/nips/ZhengC00WZL0LXZ23, DBLP:journals/corr/abs-2503-05244} to measure the quality of model-generated responses. 
There are two evaluation approaches including pointwise grading and pairwise comparison. Though widely adopted due to its simplicity, pointwise grading exhibits limited discriminative capabilities and relatively high variance. On the contrary, pairwise comparison evaluates the response against a high-quality reference, capturing the subtle differences and potential direction of improvement. 
By providing more discriminative reward signals, pairwise comparison incentivizes the model to generate a better response and defeat high-quality references for positive rewards. 
Therefore, our reward design is as follows:

\vspace{-2mm}
{\small
\[
r_{\text{quality}}(\mathbf{x}) = 
\begin{cases}
1 & \text{if } \mathrm{Judge}(\mathbf{ref}, \mathbf{x}) = \mathbf{x} \succ \mathbf{ref} \\
0.5 & \text{if } \mathrm{Judge}(\mathbf{ref}, \mathbf{x}) = \mathbf{x} \equiv \mathbf{ref} \\
0 & \text{if } \mathrm{Judge}(\mathbf{ref}, \mathbf{x}) = \mathbf{x} \prec \mathbf{ref}
\end{cases}
\]
}
\vspace{-2mm}

\noindent where $r_{\text{quality}}(\mathbf{x})$ denotes the reward for a generated response $\mathbf{x}$; $\mathbf{ref}$ represents the high-quality reference response; and $\mathrm{Judge}(\mathbf{ref}, \mathbf{x})$ is the evaluation function performed by the LLM-based judge to compare $\mathbf{x}$ with $\mathbf{ref}$.

In our experiments detailed in Appendix~\ref{appendix:model_agreement}, our model judge achieves a relatively high agreement (0.75) with human judges. Furthermore, LLM judges are known to exhibit position bias~\citep{DBLP:conf/nips/ZhengC00WZL0LXZ23} in pairwise comparisons, systematically favoring the first response. 
To impose additional learning pressure, we deliberately place the model-generated response in the second position, thereby introducing \textit{positional disadvantage} in training. 
This avoids the need for position-swapped comparisons and halves the evaluation cost, while encouraging the model to generate stronger outputs from a less favorable position.

\begin{algorithm*}[t]
    \caption{Dynamic Reference Scheduling for Long-form Writing}\label{alg:dcrf}
    \small
    \begin{algorithmic}[1]
        \State \textbf{Pre-processing:} For each instruction $w\!\in\!W$, apply Margin-aware Data Selection (Section~\ref{sec:data_selection}) to obtain a stage-wise reference list $\mathcal{R}^{(w)} = \{r_{\pi}^{(w)}, r_{1}^{(w)}, r_{2}^{(w)}, \dots\}$ ordered by ascending quality.
        \State \textbf{Input: } Instruction set $W$; reference lists $\{\mathcal{R}^{(w)}\}_{w\in W}$; policy model $\pi_{\theta}$; RL updater $\mathcal{A}$ (e.g., PPO); batch size $B$.
        \State Initialize reference pointer $t_w \leftarrow 1$ \textbf{for all} $w \in W$ \Comment{current reference index}
        \While{training not finished}
            \State Sample batch $\mathcal{B} = \{w_k\}_{k=1}^{B}$ from $W$
            \ForAll{$w_k \in \mathcal{B}$}
                \State $r_k \gets \mathcal{R}^{(w_k)}[t_{w_k}]$ \Comment{current reference}
                \State Generate response $g_k \gets \pi_{\theta}(w_k)$
                \State Compute reward $R_k \gets \mathrm{Judge}(r_k, g_k)$ \Comment{$1$ (win), $0.5$ (tie), $0$ (loss)}
            \EndFor
            \State Update policy $\pi_{\theta} \leftarrow \mathcal{A}\bigl(\pi_{\theta}, \{(w_k, g_k, R_k)\}_{k=1}^{B}\bigr)$
            \ForAll{$w_k \in \mathcal{B}$ \textbf{such that} $R_k = 1$} \Comment{reference surpassed}
                \If{$t_{w_k} < |\mathcal{R}^{(w_k)}|$}
                    \State $t_{w_k} \leftarrow t_{w_k} + 1$ \Comment{promote to next stronger reference}
                \EndIf
            \EndFor
        \EndWhile
    \end{algorithmic}
\end{algorithm*}

\subsection{Dynamic Reference Scheduling}
\label{sec:dynamic_curriculum}

Curriculum Learning~\citep{DBLP:conf/icml/BengioLCW09} schedules progressive task difficulty for better learning efficiency. 
Previous efforts utilize offline-calculated difficulty for scheduling~\citep{shi2025efficient, DBLP:journals/corr/abs-2503-17287} or introducing additional rollouts during training for adaptive sample selection~\citep{bae2025online, DBLP:journals/corr/abs-2503-14476}. 
Though effective in reasoning-centered RL, these methods suffer from either non-adaptive difficulty estimates or increased inference overhead.

To address this issue, we propose a \textit{Dynamic Reference Scheduling} approach that encourages the policy model to sequentially outperform references of ascending quality. 
With the algorithm detailed in Algorithm~\ref{alg:dcrf}, our framework introduces a more competitive reference when the policy model beats the current one during training, enabling asynchronous per-sample difficulty updates and dynamic adaptivity to the evolving model. 

\noindent\textbf{Prior to Training: Data Preparation.}
Given a set of writing instructions \( W \), we first apply the Margin-aware Data Selection strategy as elaborated in Sec~\ref{sec:data_selection}, obtaining multiple competitive references \(\mathcal{R} = \{r_\pi, r_1, r_2, \dots\}\) and their corresponding LLM-judged quality scores \(\mathcal{S} = \{s_\pi, s_1, s_2, \dots\}\) for each instruction. The references are then sorted in ascending order of quality to produce a stage-wise reference list \( \mathcal{R}_s = \{r_{q1}, r_{q2}, \dots\} \). To maintain sufficient positive feedback early in training, we deliberately include the response from the initial policy model \( \pi \) in the reference set, as the other reference-generation LLMs are generally more competent.

\noindent\textbf{During Training: Dynamic Scheduling. }
At the start of training, each instruction is initialized with the lowest-quality reference \( r_{q1} \), which is comparable to the initial policy model's response. As the model evolves during training, the model gradually generates higher-quality responses during rollouts and receives positive rewards in some of the pairwise comparisons. Subsequently, the defeated references \( r_{t} \) are replaced with marginally stronger ones \( r_{t+1} \) while the undefeated references are retained, progressively increasing the challenge without overwhelming the model, in alignment with the model’s evolving capability. 
This dynamic and adaptive reference update mechanism establishes an asynchronous learning schedule for each writing instruction and effectively incentivizes the model to consistently perform better. 

\section{Experiments}
\label{sec:experiments}

To demonstrate the effectiveness of Writing-RL, we conduct experiments on writing-oriented fine-tuned LLMs to see whether it can further advance long-form writing capabilities beyond SFT.

\begin{table*}
\begin{center}
\resizebox{\linewidth}{!}{
\begin{tabular}{c|cc|cccc}
\toprule
\multirow{2}{*}{Model}            & \multicolumn{2}{c|}{Writing-Oriented Training} & \multicolumn{4}{c}{Long-form Writing Evaluation} \\ 
\cmidrule{2-7} 
                                  & SFT                    & RL                    & WritingBench   & Creative-W.B.       & \multicolumn{1}{c|}{LongBench-Write} & Average        \\ 
\midrule
\rowcolor{gray!10}
\multicolumn{7}{c}{\textbf{(a) Proprietary LLMs}}\\
\cmidrule{1-7}
Qwen-Plus                         & --                     & --                    & 77.62          & 76.78          & \multicolumn{1}{c|}{95.42}           & 83.27          \\
GPT-4o                            & --                     & --                    & 83.42          & 80.45          & \multicolumn{1}{c|}{92.92}           & 85.60          \\ 
\midrule
\rowcolor{gray!10}
\multicolumn{7}{c}{\textbf{(b) Writing-Oriented Fine-Tuned LLMs}}\\
\cmidrule{1-7}
Suri-7B                           & \cmark                 & \xmark                & 49.70          & 18.44          & \multicolumn{1}{c|}{33.44}           & 33.86          \\
Longwriter-9B                     & \cmark                 & DPO                   & 79.10          & 44.15          & \multicolumn{1}{c|}{80.83}           & 68.03          \\ 
Longwriter-Zero-32B               & \cmark                 & GRPO                  & 82.92          & 61.14          & \multicolumn{1}{c|}{85.90}           & 76.65          \\ 
\midrule
\rowcolor{gray!10}
\multicolumn{7}{c}{\textbf{(c) Qwen2.5-7B-Instruct Model Family}}\\
\cmidrule{1-7}
Qwen2.5-7B-Instruct               & \xmark                 & \xmark                & 73.16          & 49.29          & \multicolumn{1}{c|}{87.20}           & 69.88          \\
Qwen2.5-7B-WritingBench-SFT (12k) & \cmark                  & \xmark                & 83.71          & 70.24          & \multicolumn{1}{c|}{92.56}           & 82.17          \\
Qwen2.5-7B-WritingBench-SFT (24k) & \cmark                  & \xmark                & 83.76          & 69.60          & \multicolumn{1}{c|}{92.65}           & 82.00          \\
Qwen2.5-7B-Reference-SFT          & \cmark                  & \xmark                & 84.23          & 68.89          & \multicolumn{1}{c|}{92.88}  & 82.00          \\
Qwen2.5-7B-Writing-RL (Ours)      & \cmark                  & PPO                   & \textbf{87.23} & 73.19 & \multicolumn{1}{c|}{\textbf{93.06}}           &   84.49 \\
Qwen2.5-7B-Writing-RL (Ours)     & \cmark    & GRPO & 86.29          & \textbf{75.41} & 91.84          & \textbf{84.51} \\

\midrule
\rowcolor{gray!10}
\multicolumn{7}{c}{\textbf{(d) Qwen2.5-14B-Instruct Model Family}}\\
\cmidrule{1-7}
Qwen2.5-14B-Instruct              & \xmark                 & \xmark                & 72.67          & 56.89          & \multicolumn{1}{c|}{89.20}           & 72.92          \\
Qwen2.5-14B-WritingBench-SFT      & \cmark                 & \xmark                & 84.95          & 79.27          & \multicolumn{1}{c|}{93.75}           & 85.99          \\
Qwen2.5-14B-Writing-RL (Ours)     & \cmark                 & PPO                   & \textbf{88.27} & \textbf{83.17} & \multicolumn{1}{c|}{\textbf{94.17}}  & \textbf{88.53} \\

\midrule
\rowcolor{gray!10}
\multicolumn{7}{c}{\textbf{(e) Llama3.1-8B-Instruct Model Family}}\\
\cmidrule{1-7}
Llama3.1-8B-Instruct              & \xmark                 & \xmark                & 66.56          & 48.83          & \multicolumn{1}{c|}{80.79}           & 65.39          \\
Llama3.1-8B-WritingBench-SFT      & \cmark                  & \xmark                & 83.75          & 77.70          & \multicolumn{1}{c|}{90.67}           & 84.04          \\
Llama3.1-8B-Reference-SFT         & \cmark                  & \xmark                & 83.98          & 76.70          & \multicolumn{1}{c|}{91.53}           & 84.07          \\
Llama3.1-8B-Writing-RL (Ours)     & \cmark                  & PPO                   & \textbf{87.11} & \textbf{82.67} & \multicolumn{1}{c|}{\textbf{92.09}}  & \textbf{87.29} \\ 
\bottomrule
\end{tabular}
}
\caption{Evaluation results of the models trained with Writing-RL, with the highest score in each model family \textbf{bold}. Notably, our trained models perform the best within their model family, competitive with the proprietary models.}
\label{tab:results}
\end{center}
\vspace{-2mm}
\end{table*}

\subsection{Datasets}

We use two carefully-constructed writing datasets primarily designed for SFT, including LongWriter training set~\citep{DBLP:journals/corr/abs-2408-07055} and WritingBench training set~\citep{DBLP:journals/corr/abs-2503-05244}. 
As detailed in Section~\ref{sec:data_selection}, we perform the \textit{Margin-aware Data Selection} on the datasets respectively. Specifically, we first generate references for each writing instruction with the initial policy model and four competent LLMs, including Qwen-Plus~\citep{DBLP:journals/corr/abs-2412-15115}, GPT-4o~\citep{DBLP:journals/corr/abs-2410-21276}, Claude-3.7~\citep{anthropic2024claude3sonnet} and DeepSeek-R1~\citep{DBLP:journals/corr/abs-2501-12948}. 
Then, we utilize a fine-tuned judge model~\citep{DBLP:journals/corr/abs-2503-05244}, which is optimized for evaluating long-form writing responses, to grade the responses in multiple dimensions.
Finally, after the selection process, we obtain 1.5k chosen samples from each dataset for further reinforcement learning. Each sample contains a writing instruction and references ordered by ascending quality. Additionally, we present analysis of topic coverage in Appendix~\ref{sec:dataset_analysis}.

\subsection{Training Setup}

To fully harness the potential of reinforcement learning, we use two writing-expert LLMs as the base models for RL, which are primarily fine-tuned with the full WritingBench training set, denoted as \textit{Qwen2.5-7B-WritingBench-SFT} and \textit{Llama3.1-8B-WritingBench-SFT} respectively. 

With the proposed Writing-RL, we use the PPO algorithm~\citep{DBLP:journals/corr/SchulmanWDRK17} to optimize the models for long-form writing. During the training process, we adopt Qwen-Plus to serve as pairwise comparison judge, providing rewards for policy optimization. We include more details about reward model choice and training parameters in Appendix~\ref{appendix:reward_model} and Appendix~\ref{sec:appendix_a} respectively.

\subsection{Benchmarks and Baselines}

To comprehensively evaluate long-form writing capabilities of LLMs, we use three established benchmarks including WritingBench~\citep{DBLP:journals/corr/abs-2503-05244}, LongBench-Write~\citep{DBLP:journals/corr/abs-2408-07055}, and Creative-Writing-Bench (EQ-Bench creative writing split)~\citep{DBLP:journals/corr/abs-2312-06281}. The benchmarks are of broad coverage and use strong judge LLMs to evaluate the quality of generated responses. We conduct three independent runs and report average performance for both model families to reduce variance. Note that the judge LLMs adopted for evaluation are diverse and different from the rewarding judge LLM used in training, mitigating the risk of overfitting particular judge preferences to ensure a fair evaluation.

Our selected baselines include strong proprietary models, instruction fine-tuned LLMs~\citep{DBLP:journals/corr/abs-2412-15115, DBLP:journals/corr/abs-2407-21783}, writing-oriented fine-tuned LLMs~\citep{DBLP:journals/corr/abs-2503-05244, DBLP:journals/corr/abs-2408-07055, DBLP:conf/emnlp/PhamSI24, DBLP:journals/corr/abs-2506-18841}, and the models supervisedly fine-tuned on our RL dataset. More evaluation details can be found in Appendix~\ref{app:benchmark_evaluation}.

\begin{table*}[t]
\begin{center}
\resizebox{0.9\linewidth}{!}{
\begin{tabular}{c|cc|cccccc}
\toprule
\multirow{2}{*}{Model}            & \multicolumn{2}{c|}{Writing-Oriented Training} & \multicolumn{6}{c}{Evaluation}                                                                  \\ \cmidrule{2-9} 
                                  & SFT                    & RL                    & Easy  & \multicolumn{1}{c|}{Hard}  & Short & Medium & \multicolumn{1}{c|}{Long} & Overall       \\ \midrule
Qwen2.5-7B-Instruct               & \xmark                 & \xmark                & 31.8  & \multicolumn{1}{c|}{28.3}  & 38.9  & \textbf{26.0}   & \multicolumn{1}{c|}{21.3} & 29.6          \\
Qwen2.5-7B-WritingBench-SFT & \cmark                 & \xmark                & 27.6  & \multicolumn{1}{c|}{27.7}  & 35.0  & 25.1   & \multicolumn{1}{c|}{20.4} & 27.6          \\
Qwen2.5-7B-Writing-RL (Ours)      & \cmark                 & PPO                   & \textbf{35.8} & \multicolumn{1}{c|}{\textbf{29.3}} & \textbf{42.1}  & 25.7   & \multicolumn{1}{c|}{\textbf{26.5}} & \textbf{31.8} \\ \midrule \midrule
Llama3.1-8B-Instruct              & \xmark                 & \xmark                & \textbf{32.3}  & \multicolumn{1}{c|}{28.9}  & 35.6  & 27.4   & \multicolumn{1}{c|}{\textbf{26.9}} & 30.2          \\
Llama3.1-8B-WritingBench-SFT      & \cmark                 & \xmark                & 29.7  & \multicolumn{1}{c|}{27.7}  & 36.7  & 23.7   & \multicolumn{1}{c|}{24.1} & 28.4          \\
Llama3.1-8B-Writing-RL (Ours)     & \cmark                 & PPO                   & 31.2  & \multicolumn{1}{c|}{\textbf{33.8}}  & \textbf{42.2}  & \textbf{29.3}   & \multicolumn{1}{c|}{24.1} & \textbf{32.8} \\ \midrule
\end{tabular}
\vspace{-2mm}
}
\end{center}
\vspace{-2mm}
\caption{\textbf{Output-to-Input Generalization: }Evaluation results of the models trained with Writing-RL on LongBench v2, demonstrating the generalization potential from long-output generation to long-input reasoning.}
\label{tab:generalization_results}
\vspace{-2mm}
\end{table*}

\subsection{Results}

\begin{figure}[t]
    \vspace{-2mm}
    \centering
    \includegraphics[width=\linewidth]{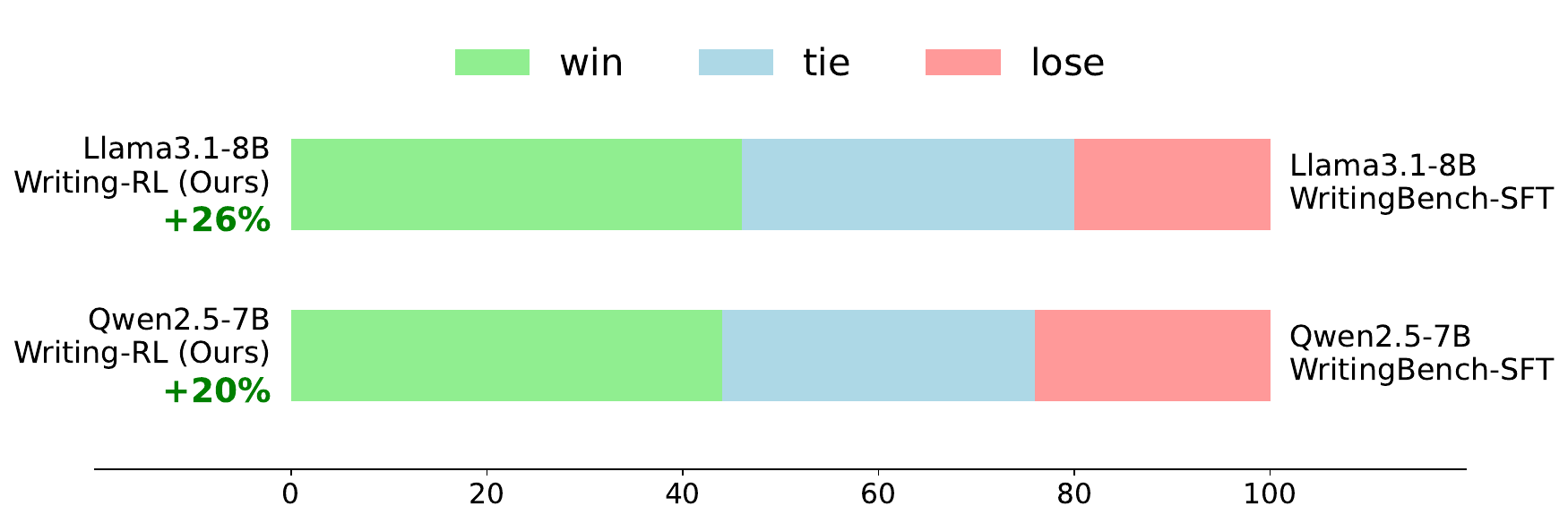}
    \vspace{-4mm}
    \caption{\textbf{Human Evaluation} results of pairwise comparison between our RL-trained models and the best-performing SFT-trained competitors.}
    \label{fig:human_eval}
    \vspace{-2mm}
\end{figure}

\noindent\textbf{Remarkable Writing Performance Gains.} As shown in Table~\ref{tab:results}, the results demonstrate that our models outperform others across the benchmarks. 
Specifically, \textit{Llama3.1-8B-Writing-RL} achieves the highest average score of 87.29, and \textit{Qwen2.5-7B-Writing-RL} follows with an average of 84.49, both showing strong performance at their scale. 
Furthermore, our approach scales effectively to larger models, yielding substantial improvements as evidenced by \textit{Qwen2.5-14B-Writing-RL} achieving an outstanding average score of 88.53.
Notably, our trained models exhibit long-form writing capabilities that match or even surpass those of proprietary models, positioning them as strong open-source alternatives for long-form generation.

\noindent\textbf{SFT Encounters Data Saturation.} We observe distinct performance trends when applying RL and SFT to relatively strong models. 

Despite using identically constructed datasets, we observe a marginal performance regression in \textit{Qwen2.5-7B-WritingBench-SFT (24k)} relative to its counterpart \textit{Qwen2.5-7B-WritingBench-SFT (12k)}.
Furthermore, the models continuously fine-tuned with high-quality references in the RL dataset, namely \textit{Llama3.1-8B-Reference-SFT} and \textit{Qwen2.5-7B-Reference-SFT}, also show minimal performance gain, or even slight degradation.
This observation potentially underscores the phenomenon of data saturation, where as the model gets stronger, it is harder to construct SFT data to effectively further improve model capability. 

\noindent\textbf{RL Leads to Further Improvements.} In contrast, models trained by Writing-RL demonstrate consistent improvements and thereby indicate the promising potential of RL to further advance model capability where SFT encounters limitations.

\noindent\textbf{Robustness Across RL Algorithms.} Applied to the Qwen2.5-7B backbone, the GRPO variant yields an average score of 84.51, performing comparably to PPO. GRPO shows a slight edge in creative writing, whereas PPO leads marginally on WritingBench and LongBench-Write. While minor task-specific variances exist, the overall parity confirms that the performance gains are fundamentally driven by data quality, reward design, and curriculum strategy, rendering the framework highly effective regardless of the underlying RL optimizer.

\vspace{-1mm}
\subsection{Human Evaluation}
\vspace{-1mm}

Furthermore, we recognize that human evaluation can complement automatic LLM-as-Judge. Therefore, we also conduct human evaluation experiments to further validate our model performance. We randomly sample 100 writing instructions in total from our evaluation datasets and generate responses using our RL-trained models and the most competitive baselines. Then, the annotators select the better-quality response under the same writing instruction. As shown in Fig.~\ref{fig:human_eval}, the results demonstrate higher win rates of our trained models, indicating their stronger long-form writing capability and better alignment with human preferences. Additionally, we include a case study of realistic writing scenarios in Appendix~\ref{sec:creative_writing_case}.

\begin{table*}[ht]
\begin{minipage}{0.48\linewidth}
\vspace{-2mm}
\begin{center}
\small
\resizebox{\linewidth}{!}{
\begin{tabular}{c|cc|c}
\toprule
\multirow{2}{*}{\makecell{Selection\\Strategy}} & 
\multirow{2}{*}{\makecell{Initial\\Score}} & 
\multirow{2}{*}{\makecell{Learning\\Potential}} & 
\multirow{2}{*}{\makecell{Writ.\\Score}} \\ 
 & & & \\ \midrule

Baseline (w/o RL)      & --    & --    & 83.71$_{\pm 0.10}$ \\
Full (w/o Selection)   & 84.20 & 3.64  & 85.91$_{\pm 0.24}$ \\
Difficulty-prioritized & 77.61 & 8.18  & 86.29$_{\pm 0.28}$ \\
Margin-aware (Ours) & 78.84 & 9.16 & \textbf{86.98}$_{\pm 0.23}$ \\
\bottomrule
\end{tabular}
}
\end{center}
\vspace{-3mm}
\caption{Comparison of different data selection strategies, indicating the benefits of higher learning potential.}
\label{ablation:data_selection}

\end{minipage}\hfill
\begin{minipage}{0.48\linewidth}
\vspace{-4mm}
\begin{center}
\small
\resizebox{\linewidth}{!}{
\begin{tabular}{c|cc|c}
\toprule
\makecell{Reward\\Strategy} & 
\makecell{Multi\\Dimension} & 
\makecell{Reference\\Based} & 
\makecell{Writ.\\Score} \\
\midrule
Baseline (w/o RL)   & \xmark & \xmark & 83.71$_{\pm 0.10}$ \\
Pointwise           & \cmark & \xmark & 84.50$_{\pm 0.09}$ \\
Pairwise (Ours)     & \cmark & \cmark & \textbf{86.98}$_{\pm 0.23}$ \\ 
\bottomrule
\end{tabular}
}
\end{center}
\vspace{-2.5mm}
\caption{Comparison of different reward designs, indicating the effectiveness of multi-dimensional pairwise LLM judges during training.}
\label{ablation:reward}
\vspace{-2mm}
\end{minipage}
\end{table*}

\begin{table*}[ht]
\begin{minipage}{0.48\linewidth}
\begin{center}
\vspace{-1mm}
\small
\resizebox{\linewidth}{!}{%
\begin{tabular}{c|ccc|c}
\toprule
\makecell{Curriculum\\Strategy}           & \makecell{Writ.\\Bench} & \makecell{C.W.\\Bench} & \makecell{Long.\\Write} & \makecell{Average\\Score}        \\ \midrule

Baseline (w/o RL)      & 83.71	&  70.24	&   92.56	&   82.17$_{\pm 0.10}$          \\
None                   & 86.51		&  71.47	&  	92.26	&  	83.41$_{\pm 0.23}$   \\
Static                 & 86.40		&  72.84		&  92.26		&  83.83$_{\pm 0.54}$      \\
Dynamic (Ours)         & \textbf{87.23}        & \textbf{73.19} & \textbf{93.06} & \textbf{84.49}$_{\pm 0.37}$ \\ 
\bottomrule
\end{tabular}
}
\end{center}
\vspace{-2mm}
\caption{Ablation of our adaptive curriculum scheduling approach, demonstrating the superiority of our method and the effectiveness of curriculum design in RL.}
\label{ablation:curriculum}
\vspace{-5mm}
\end{minipage}\hfill
\begin{minipage}{0.48\linewidth}
\vspace{-1mm}
\begin{center}
\vspace{-1mm}
\small
\resizebox{\linewidth}{!}{%
\begin{tabular}{c|cc}
\toprule
Model       & MT-Bench            & MMLU             \\ \midrule

Qwen2.5-7B-Instruct           & 7.21                & \textbf{71.85}            \\
Qwen2.5-7B-WritingBench-SFT   & 7.34                & 69.66            \\
Qwen2.5-7B-Writing-RL (Ours)  & \textbf{7.62}                & 69.75            \\ \midrule

Llama3.1-8B-Instruct          & 6.16                & \textbf{66.03}            \\
Llama3.1-8B-WritingBench-SFT  & \textbf{6.42}                & 64.98            \\
Llama3.1-8B-Writing-RL (Ours) & 6.29                & 65.02            \\ 
\bottomrule
\end{tabular}
}
\end{center}
\vspace{-1mm}
\caption{Evaluation results on the general-capabilities benchmarks MT-Bench and MMLU.}
\label{discussion:general}
\vspace{-2mm}
\end{minipage}
\vspace{-2mm}
\end{table*}
\vspace{-2mm}

\section{Generalization from Output to Input}

\begin{figure}[ht]
    \vspace{-2mm}
    \centering
    \includegraphics[width=\linewidth]{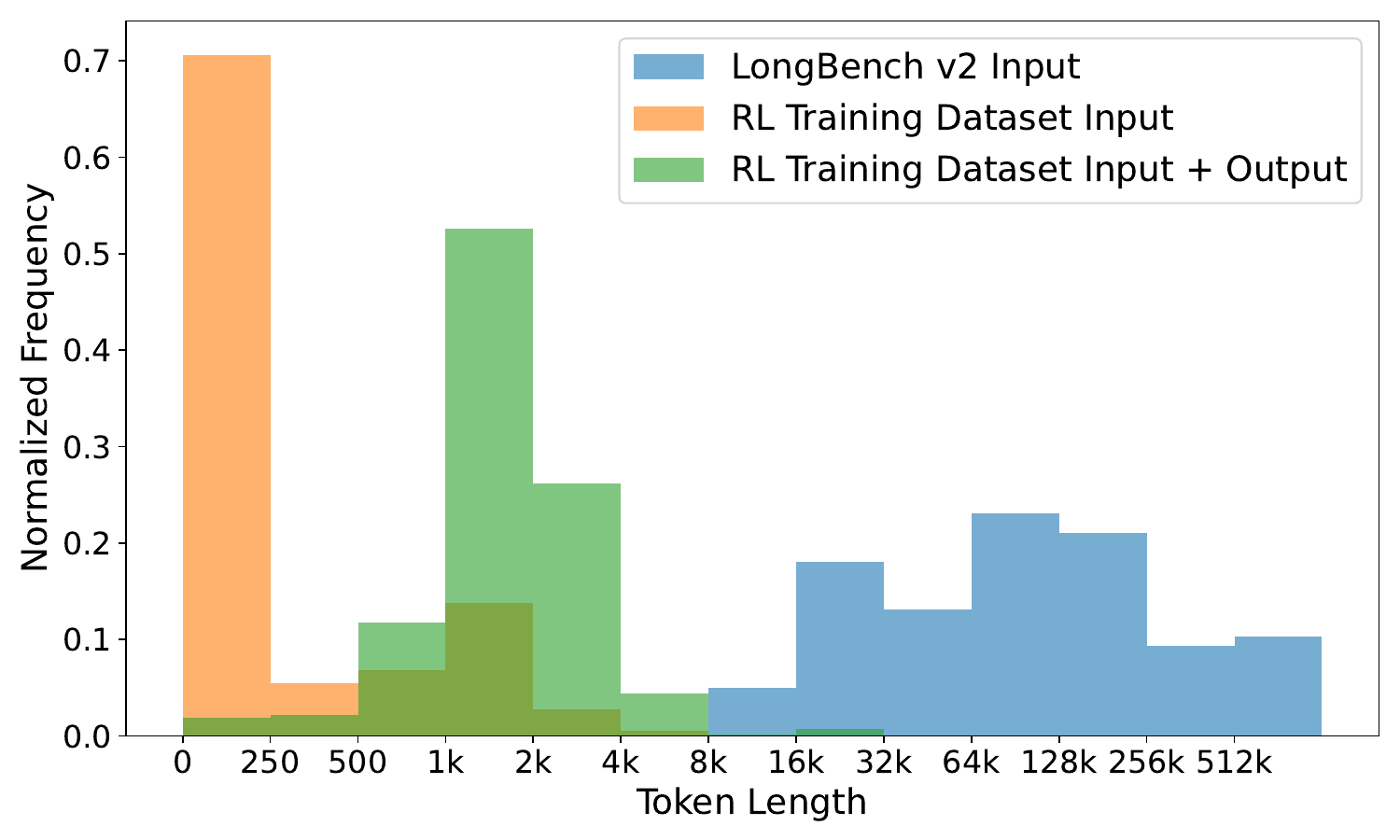}
    \vspace{-6mm}
    \caption{Length distribution of our \textit{long-output} RL training dataset and the \textit{long-input} evaluation dataset LongBench v2.}
    \label{fig:length_dis}
    \vspace{-4mm}
\end{figure}

To understand the influence of Writing-RL on long-context capabilities of Writing-RL, we adopt the challenging long-context reasoning benchmark LongBench v2~\citep{DBLP:journals/corr/abs-2412-15204} to evaluate long-input reasoning. Notably, as shown in Figure~\ref{fig:length_dis}, the input lengths in LongBench v2 are substantially longer than those in our training set, mostly exceeding not only the input lengths but also the total input–output lengths.

As detailed in Table~\ref{tab:generalization_results}, our findings are encouraging. 
Beyond improved performance in long-form generation, the writer models fine-tuned with Writing-RL also exhibit surprising generalization to long-context reasoning with substantially longer inputs, while the SFT-trained counterparts show slight performance degradation in this regime. To further understand and analyze this interesting phenomenon, we give an intuitive explanation to the following research questions. 

\noindent\textit{\textbf{Why does long-output training generalize to long-input reasoning?}} 
Generating high-quality long-form text inherently requires a deep and holistic understanding of the preceding context.
Therefore, long-generation RL encourages LLMs to develop long-input understanding capabilities as a prerequisite for producing coherent long-outputs.

\noindent\textit{\textbf{What is the source of the generalization phenomenon?}} 
We systematically ablate key components of our pipeline to isolate the drivers of generalization on LongBench-v2, revealing that the RL training paradigm itself serves as the primary catalyst for performance gains. 
While continuing SFT on the RL dataset slightly degrades the base model's capabilities, employing RL strategies consistently drives positive transfer. 
We include more controlled experiments, data analysis and case study in Appendix~\ref{appendix:long-generalization}. 

\noindent\textit{\textbf{Why does long-output RL generalize better than SFT?}} SFT forces the model to imitate and memorize the behaviors of the training samples, while RL aligns model behavior with outcome-based objectives via reward signals. Therefore, by empowering the model to enhance its underlying capabilities, RL generalizes better. This observation is also consistent with recent findings in other domains~\citep{DBLP:journals/corr/abs-2501-17161, shen2025vlm}.

\noindent\textit{\textbf{How might these findings inform long-context training?}}
The generalization from long-output generation to long-input reasoning may suggest a mutually beneficial relationship between long-input and long-output training. Integrating both perspectives may lead to more effective long-context training strategies, and we leave the systematic exploration of this promising approach to future work.

\section{Discussions}

\vspace{-1mm}
\subsection{Analysis of Data Selection Strategy} 
\vspace{-1mm}
\label{sec:ablation_1}

To validate our Margin-aware Data Selection strategy, we conduct experiments on the WritingBench training set, training \textit{Qwen2.5-7B-WritingBench-SFT} model with high-quality references generated by Qwen-Plus.
We adopt WritingBench to benchmark writing capability due to its broad coverage and efficiency, and we report mean performance and standard deviation across three independent evaluation runs.
As shown in Table~\ref{ablation:data_selection}, the results indicate that our strategy can boost learning efficiency by choosing samples with higher learning potential, highlighting the effectiveness of \textit{learning potential} as the key metric.

\vspace{-1mm}
\subsection{Analysis of Reward Design}
\label{sec:ablation_reward}
\vspace{-1mm}

To validate our pairwise reward design, we compare our reward mechanism with the widely adopted pointwise grading method~\citep{DBLP:conf/nips/ZhengC00WZL0LXZ23, liu2025inferencetimescalinggeneralistreward}, which utilizes LLM-based judge to provide a scalar rating representing response quality. 
We follow the experiment setting in Section~\ref{sec:ablation_1}.
The results shown in Table~\ref{ablation:reward} demonstrate the superiority of our approach in providing more discriminative rewards, incentivizing the model to further advance writing capabilities.

\begin{table*}[ht]
    \centering
    \resizebox{\linewidth}{!}{%
    \begin{tabular}{llcccc}
        \toprule
        \textbf{Model} & \textbf{Judge Model} & \textbf{WritingBench} & \textbf{Creative} & \textbf{LongBench-Write} & \textbf{Average} \\
        \midrule
        Qwen2.5-7B-WritingBench-SFT & -- &    83.71	& 70.24	& 92.56	& 82.17   \\
        \midrule
        Qwen2.5-7B-Writing-RL (V3 Judge) & DeepSeek-V3 & \textbf{87.87} & 72.66 & 92.36 & 84.30 \\
        Qwen2.5-7B-Writing-RL (Ours) & Qwen-Plus & 87.23 & \textbf{73.19} & \textbf{93.06} & \textbf{84.49} \\
        \bottomrule
    \end{tabular}%
    \vspace{-2mm}
    }
    \vspace{-2mm}
    \caption{Performance comparison using different judge models for reward calculation. Replacing Qwen-Plus with DeepSeek-V3 yields comparable gains over the SFT baseline, demonstrating that our method's effectiveness is robust to the choice of the reward model.}
    \label{tab:judge_ablation}
\vspace{-2mm}
\end{table*}

\begin{table}
\vspace{-2mm}
\begin{center}
\small
\begin{tabular}{c|c}
\toprule
Reference Quality           &  Score \\ \midrule
Self-Generated &            86.80                 \\
Qwen-Plus                   & \textbf{86.98}              \\
DeepSeek-R1                 & 86.15                  \\
Best Reference              & 82.51              \\ \bottomrule
\end{tabular}
\end{center}
\vspace{-2mm}
\caption{Comparison of different reference quality settings.}
\label{ablation:quality}
\vspace{-2mm}
\end{table}

\vspace{-1mm}
\subsection{Analysis of reference quality}
\label{sec:ablation_reference}
\vspace{-1mm}

Under the Pairwise Comparison Reward Mechanism, reference quality dictates the difficulty of obtaining positive rewards, directly impacting training stability and performance. To evaluate this, we test static reference sets generated by various LLMs, alongside a combined \textit{Best Reference} set and a \textit{Self-Generated} baseline.

As Table~\ref{ablation:quality} demonstrates, reference quality is critical for effective optimization. 
Low-quality references (\textit{e.g., Self-Generated}) cause rapid reward saturation, halting progress once the model easily achieves high win rates. 
Conversely, overly high-quality references (\textit{e.g., Best Reference}) create sparse positive rewards early in training, destabilizing the learning process. 
These findings highlight a fundamental limitation of static reference scheduling: its inability to adapt to the policy model's evolving capabilities.

\vspace{-1mm}
\subsection{Ablation on Curriculum Scheduling}
\vspace{-1mm}

Given the importance of reference quality and the limitations of fixed references discussed in Section~\ref{sec:ablation_reference}, we propose \textit{Dynamic Reference Scheduling}, which progressively introduces higher-quality references as the model evolves. 
To validate this strategy, we conduct an ablation study comparing three curriculum setups: mixed training without scheduling, static scheduling which partitions the dataset into two subsets with different difficulty, and our proposed dynamic scheduling. 
As shown in Table~\ref{ablation:curriculum}, the results confirm the superiority of our approach. 
Furthermore, both static and dynamic scheduling outperform the no-curriculum baseline, demonstrating the effectiveness of incorporating curriculum into RL.

\vspace{-1mm}
\subsection{Ablation of Judge Model}
\label{sec:judge_ablation}
\vspace{-1mm}

To verify that the performance improvements observed in our Writing-RL framework are not dependent on the specific preference distribution or biases of a single judge model, we conducted an ablation study replacing our default judge, Qwen-Plus, with DeepSeek-V3.

As illustrated in Table~\ref{tab:judge_ablation}, both RL-trained variants notably outperform the SFT baseline. The model trained with DeepSeek-V3 as the judge achieves an average score of 84.30, which is statistically comparable to our default Qwen-Plus judge. This consistency confirms that the effectiveness of our iterative RL pipeline is driven by the Writing-RL module design rather than specific judge model.

\vspace{-1mm}
\subsection{Influence on General Capabilities}
\vspace{-1mm}

The evaluation results presented in Table \ref{discussion:general} provide insights into the impact of writing-oriented training on the general capabilities of LLMs, as assessed by the MMLU~\citep{DBLP:conf/iclr/HendrycksBBZMSS21} and MT-Bench~\citep{DBLP:conf/nips/ZhengC00WZL0LXZ23}. 
MMLU results indicate that RL-trained models maintain parity with their SFT counterparts, evincing negligible degradation in core knowledge. Furthermore, long-output variants surpass baseline instruct models on MT-Bench. Collectively, these findings demonstrate that our RL framework enhances long-form generation while preserving general capabilities.

\vspace{-1mm}
\subsection{Cost Analysis}
\vspace{-1mm}

To assess the feasibility of Writing-RL, we provide a comprehensive cost analysis in Appendix~\ref{sec:cost_analysis}. 
Our breakdown reveals that the primary monetary expense is a one-time, upfront cost for dataset construction (approximately \$$609$ in API fees and $76$ GPU hours). 
In contrast, the recurring cost for each Reinforcement Learning training run remains highly accessible, requiring roughly \$$54$ in API usage for the reward judge and $24 \times 8$ GPU hours. 
This demonstrates that once the initial high-quality dataset is curated, our approach is both highly cost-effective and scalable for future adaptations.

\section{Conclusion}

In this work, we propose \textbf{Writing-RL}: an \textit{Adaptive Curriculum Reinforcement Learning} framework, which consists of \textit{Margin-aware Data Selection}, \textit{Pairwise Comparison Reward} and \textit{Dynamic Reference Scheduling}. Our experiments demonstrate its effectiveness on enhancing long-form writing capabilities and the performance gain successfully generalizes from long-output generation to long-input reasoning, indicating a promising perspective for long-context training.

\section*{Limitations}
Here we discuss several limitations of this work.

\noindent\textbf{To scale up model size.} While the performance gain by training writer models at 7B and 14B scales is relatively large, there remains considerable room for exploration at larger model scales. 
Prior research has shown that the underlying capability of the base model plays a crucial role in the effectiveness of RL~\citep{DBLP:journals/corr/abs-2503-01307}. 
Therefore, applying Writing-RL to stronger models may lead to even greater performance improvements, as well as more pronounced generalization effects from long-output generation to long-input reasoning.

\noindent\textbf{To explore the \textit{zero phenomenon} of RL.} This work demonstrates that reinforcement learning, when applied to long-form generation, can elicit strong performance gains and even induce generalization to long-input reasoning. While an intriguing research direction is to investigate this phenomenon from a more fundamental perspective by directly applying RL to base models without prior supervised fine-tuning. Such a setup may offer clearer insight into whether RL alone is sufficient to induce strong long-form generation capabilities.

\section*{Acknowledgements}

This work is supported by the National Natural Science Foundation of China (No. 62276152, 62236011), Key Laboratory of Ethnic Language Intelligent Analysis and Security Governance of MOE, Minzu University of China, Beijing, China.

\bibliography{latex/custom}

\newpage
\appendix

\section{Implementation and Training Settings}
\label{sec:appendix_a}

\subsection{Implementation Details}
In this section, we introduce the implementation details of our proposed RL framework.

\noindent\textbf{Margin-aware Data Selection.\label{app:data_selection}} We use several close-sourced LLMs to generate high-quality references for further training, including Qwen-plus~\citep{DBLP:journals/corr/abs-2412-15115}, GPT-4o~\citep{DBLP:journals/corr/abs-2410-21276}, Claude 3.7~\citep{anthropic2024claude3sonnet} and DeepSeek-R1~\citep{DBLP:journals/corr/abs-2501-12948}. We set the inference temperature to 0.1 for balanced diversity and quality, and we remain other parameters to the default setting.

In our pointwise grading process, we utilize the state-of-the-art evaluation procedure proposed by WritingBench~\citep{DBLP:journals/corr/abs-2503-05244}, which includes generating sample-dependent evaluation criteria, then uses a fine-tuned LLM to grade the answers from multiple dimensions, finally averages the dimensional scores to give a scalar rating. 
We use Qwen-Plus~\citep{DBLP:journals/corr/abs-2412-15115} to generate the evaluation dimensions and we use the same evaluation prompt as WritingBench~\citep{DBLP:journals/corr/abs-2503-05244} for the Judge Model. 

\begin{tcolorbox}[breakable, colback=bg_rl, colframe=border_gray, title=\textbf{Evaluation Prompt Template}, fonttitle=\bfseries\small, boxrule=0.5mm]
Evaluate the Response based on the Query and criteria provided.\\
\\
** Criteria **\\
```\{criteria\}```\\
\\
** Query **\\
```\{query\}```\\
\\
** Response **\\
```\{response\}```\\
\\
Provide your evaluation based on the criteria:\\
\\
```\{criteria\}```\\
\\
Provide reasons for each score, indicating where and why any strengths or deficiencies occur within the Response. Reference specific passages or elements from the text to support your justification.\\
Ensure that each reason is concrete, with explicit references to the text that aligns with the criteria requirements.\\
\\
Scoring Range: Assign an integer score between 1 to 10\\
\\
** Output format **\\
Return the results in the following JSON format, Only output this JSON format and nothing else:\\
```json\\
\{\{\\
    "score": an integer score between 1 to 10,\\
    "reason": "Specific and detailed justification for the score using text elements."\\
\}\}
```
\end{tcolorbox}

\noindent\textbf{Pairwise Comparison Reward Mechanism.}

We use the Qwen-Plus~\citep{DBLP:journals/corr/abs-2412-15115} model to judge the quality of the generated responses. The pairwise comparison prompts used in our experiment are adapted from ~\cite{DBLP:conf/nips/ZhengC00WZL0LXZ23}
and ~\cite{DBLP:journals/corr/abs-2503-05244}.

For the training samples in LongWriter~\citep{DBLP:journals/corr/abs-2408-07055}
dataset, we use the original evaluation dimensions and the prompt is as follows.

\begin{tcolorbox}[breakable, colback=bg_rl, colframe=border_gray, title=\textbf{Default Pairwise Comparison Prompt}, fonttitle=\bfseries\small, boxrule=0.5mm]
Please act as an impartial judge and evaluate the quality of the responses provided by two AI assistants to the user question displayed below. You should choose the assistant that follows the user's instructions and answers the user's question better. Your evaluation should consider factors such as the helpfulness, relevance, accuracy, depth, creativity, and level of detail of their responses. Begin your evaluation by comparing the two responses and provide a short explanation. Avoid any position biases and ensure that the order in which the responses were presented does not influence your decision. Do not allow the length of the responses to influence your evaluation. Do not favor certain names of the assistants. Be as objective as possible. After providing your explanation, output your final verdict by strictly following this format: "[[A]]" if assistant A is better, "[[B]]" if assistant B is better, and "[[C]]" for a tie. NOTE: If the response contains severe repetition or redundancy, it should be viewed as low quality score, losing the comparison. \\
\\
{User Question} \\
\{{question}\} \\

{The Start of Assistant A's Answer} \\
\{{answer\_a}\} \\
{The End of Assistant A's Answer} \\

{The Start of Assistant B's Answer} \\
\{{answer\_b}\} \\
{The End of Assistant B's Answer}
\end{tcolorbox}

For the training samples in WritingBench~\citep{DBLP:journals/corr/abs-2503-05244} training dataset, we use the generated criteria as the original paper recommends and the prompt is as follows.

\begin{tcolorbox}[breakable, colback=bg_rl, colframe=border_gray, title=\textbf{Criteria Pairwise Comparison Prompt}, fonttitle=\bfseries\small, boxrule=0.5mm]
Please act as an impartial judge and evaluate the quality of the responses provided by two AI assistants to the user question displayed below. You should choose the assistant that follows the user's instructions and answers the user's question better. Your evaluation should consider the following dimensions.

{criteria}

Begin your evaluation by comparing the two responses and provide a short explanation. Avoid any position biases and ensure that the order in which the responses were presented does not influence your decision. Do not allow the length of the responses to influence your evaluation. Do not favor certain names of the assistants. Be as objective as possible. After providing your explanation, output your final verdict by strictly following this format: "[[A]]" if assistant A is better, "[[B]]" if assistant B is better, and "[[C]]" for a tie. NOTE: If the response contains severe repetition or redundancy, it should be viewed as low quality score, losing the comparison. \\
\\
{User Question} \\
\{{question}\} \\

{The Start of Assistant A's Answer} \\
\{{answer\_a}\} \\
{The End of Assistant A's Answer} \\

{The Start of Assistant B's Answer} \\
\{{answer\_b}\} \\
{The End of Assistant B's Answer}
\end{tcolorbox}

\subsection{Training Parameters}

We display the key training parameters used in our training experiments. We adopt the effective reinforcement training framework VeRL~\citep{sheng2024hybridflow} to train our models. 
In our experiment, we use the proximal policy optimization (PPO)~\citep{DBLP:journals/corr/SchulmanWDRK17} algorithm with generalized advantage estimation (GAE) as the advantage estimator. In the parameters about GAE, we set $\gamma = 1.0$ and $\lambda = 1.0$.
The critic model shares the same backbone as the policy model and is initialized with identical parameters.
As for the learning rate, we set a linear warm-up phase to stablize training: a low learning rate ($1\times 10^{-6}$) with a warm-up ratio of $0.4$ for the actor model, while the critic adopts a higher learning rate ($1\times 10^{-5}$) with a warm-up ratio of $0.05$.

The training process is conducted using a batch size of $32$ for training, with a maximum prompt length of $4,096$ tokens and response length capped at $10,000$ tokens to accommodate long-form generation tasks. 
We enable the parameter/optimizer offloading via Fully Sharded Data Parallel (FSDP) to support efficient multi-GPU training and the training is conducted on 8 GPUs. 
We utilize a rollout strategy based on the vLLM engine with a tensor model parallel size of $2$ and the sampling temperature during rollout is set to $1.0$ throughout training to promote exploration.
The KL divergence penalty is set to a modest coefficient of $0.001$. 
We train each model for about $400$ steps and evaluate the checkpoints on the validation set each $50$ steps.

In terms of the parameters about the SFT stage, the details are documented in \cite{DBLP:journals/corr/abs-2503-05244} and we record the key parameters here. 
The models are trained using the AdamW optimizer with $\beta_1 = 0.9$, $\beta_2 = 0.999$, and $\epsilon = 1 \times 10^{-8}$. The base learning rate was $7 \times 10^{-6}$, and a cosine decay scheduler with a warmup ratio of $0.1$ (i.e., 10\% of total training steps) was applied. Training runs for $5$ epochs with a per-device batch size of $1$ and gradient accumulation over $4$ steps, yielding an effective batch size of $128$ across 32 GPUs.

\subsection{Reward model choice}
\label{appendix:reward_model}

\begin{table*}
\begin{center}
\small
\begin{tabular}{c|ccc}
\toprule
Model & Agreement & Cost (Input / Output, \$/M tokens) & First Token Latency (s) \\
\midrule
Claude-3.7-Sonnet & 0.82 & 3.0 / 15.0 & 5.35 \\
R1                & 0.76 & --         & --   \\
GPT-4o (2024-11-20) & 0.70 & 2.5 / 10.0 & 2.19 \\
Qwen-Plus         & 0.75 & 0.4 / 1.2  & 1.16 \\
\bottomrule
\end{tabular}
\end{center}
\caption{Performance and cost comparison of different LLM judges.}
\label{appendix:reward_model_comparison}
\end{table*}

To select an appropriate model to serve as the pairwise judge during training, we analyze the human agreement, cost and latency of several cutting-edge LLMs. As shown in Table~\ref{appendix:reward_model_comparison}, Qwen-plus has already achieved a high agreement with human judges, demonstrating its reward-giving capablities and making it a reliable choice for the training writer models. As shown in the following human evaluation results, qwen-plus has reached a remarkable agreement of 0.75, on par with R1 and surpassing gpt-4o-2024-11-20. Furthermore, GPT-4o and Claude models are widely adopted as judges in LLM benchmarks. If we use GPT series as training-time judges, the evaluation will be biased and unreliable. Therefore, we use a different training-time judge rather than the test-time judges.

RL requires a large amount of pairwise rewarding, therefore leading to huge API costs and high efficiency demands. As shown in the following results, qwen-plus has a remarkably lower price than gpt-4o and claude-3.7-sonnet and possesses the lowest first token latency.

\subsection{Cost Analysis}
\label{sec:cost_analysis}

To facilitate community reproduction and assess the feasibility of our approach, we provide a comprehensive breakdown of the monetary and computational costs involved. Our analysis distinguishes between the one-time offline costs of data curation (Stage 1) and the recurring costs of Reinforcement Learning (RL) training (Stage 2).

\subsubsection{Stage 1: Dataset Construction (One-Time Cost)}

The data selection stage involves generating competitive reference responses and scoring candidates. We utilized a mix of closed-source APIs and open-source models. As detailed in Table~\ref{tab:stage1_cost}, the total API cost for generating high-quality reference responses was \$608.71, utilizing Claude-3.7-Sonnet, GPT-4o, Qwen-Plus, and DeepSeek-R1. 

We note that the majority of this expense (\$375.52) was attributed to Claude-3.7-Sonnet. Future reproductions can significantly reduce costs by substituting this with more cost-effective alternatives identified in our survey. Additionally, inference with the policy model and local scoring required 34.51 and 41.25 GPU hours, respectively. 

Crucially, \textbf{these are one-time offline costs used to produce a reusable, high-quality dataset.} Once constructed, this dataset eliminates these expenses for all subsequent training runs and community adaptations.

\begin{table*}[ht]
    \small
    \centering
    \begin{tabular}{lcccc}
        \toprule
        \textbf{Model} & \textbf{Input Price} & \textbf{Output Price} & \textbf{Total Cost} & \textbf{GPU Hours} \\
         & (\$/1M) & (\$/1M) & (\$) & (GPU) \\
        \midrule
        Claude-3.7-Sonnet & 3.00 & 15.00 & 375.52 & -- \\
        GPT-4o & 2.50 & 10.00 & 164.33 & -- \\
        Qwen-Plus & 0.40 & 1.20 & 27.74 & -- \\
        DeepSeek-R1 & 0.57 & 2.29 & 31.12 & -- \\
        Qwen2.5-7B-Instruct (SFT) & -- & -- & -- & 34.51 \\
        Qwen2.5-7B-Instruct (Judge) & -- & -- & -- & 41.25 \\
        \midrule
        \textbf{Total} & & & \textbf{608.71} & \textbf{75.76} \\
        \bottomrule
    \end{tabular}
    \caption{Cost breakdown for Stage 1: Data Selection.}
    \label{tab:stage1_cost}
\end{table*}

\subsubsection{Stage 2: RL Training}

For the RL training process, costs are recurring per run. We employed Qwen-Plus as the external judge model for reward assignment. As shown in Table~\ref{tab:stage2_cost}, a complete training run incurs approximately \$53.50 in API costs and requires 24 hours on 8 GPUs.

\begin{table*}[ht]
    \small
    \centering
    \begin{tabular}{lcccc}
        \toprule
        \textbf{Component} & \textbf{Input Price} & \textbf{Output Price} & \textbf{Run Cost} & \textbf{Compute Time} \\
         & (\$/1M) & (\$/1M) & (\$) & (8$\times$GPUs) \\
        \midrule
        Qwen-Plus API (Judge) & 0.40 & 1.20 & 53.50 & -- \\
        Model Training & -- & -- & -- & 24 Hours \\
        \bottomrule
    \end{tabular}
    \caption{Cost breakdown for Stage 2: RL Training.}
    \label{tab:stage2_cost}
\end{table*}

To ensure cost-efficiency without compromising quality, we conducted a comparative survey of potential judge models (Table~\ref{appendix:reward_model_comparison}). Qwen-Plus was selected because it achieves a high agreement rate with human judges (0.75), comparable to DeepSeek-R1 (0.76), while offering significantly lower latency (1.16s first token latency) and reduced API costs compared to GPT-4o and Claude-3.7-Sonnet.

In conclusion, the primary monetary cost is front-loaded in the dataset construction phase. The per-run training cost remains moderate (\$53.50 + 24 compute hours), which we believe is a reasonable expenditure given the performance gains and the reusability of the constructed resources.

\section{Benchmarks and Evaluation Methods\label{app:benchmark_evaluation}} 

In this section, we introduce the benchmarks and evaluation prompt templates used in our experiments.

\paragraph{LongBench-Write}
LongBench-Write~\citep{DBLP:journals/corr/abs-2408-07055} is designed to evaluate the LLM long-form generation abilities, which focuses on generating coherent outputs exceeding 10000 words, addressing challenges in maintaining consistency and quality over extended text.
Key evaluation metrics include coherence, fluency and topic relevance.
In this work, we use the Quality Score as the metric. 
The evaluation prompt template used is as follows:

\begin{tcolorbox}[breakable, colback=bg_rl, colframe=border_gray, title=\textbf{Evaluation Prompt Template}, fonttitle=\bfseries\small, boxrule=0.5mm]
You are an expert in evaluating text quality. Please evaluate the quality of an AI assistant's response to a user's writing request. Be as strict as possible.

You need to evaluate across the following six dimensions, with scores ranging from 1 to 5. The scoring criteria from 5 to 1 for each dimension are as follows:

1. Relevance: From content highly relevant and fully applicable to the user's request to completely irrelevant or inapplicable.

2. Accuracy: From content completely accurate with no factual errors or misleading information to content with numerous errors and highly misleading.

3. Coherence: From clear structure with smooth logical connections to disorganized structure with no coherence.

4. Clarity: From clear language, rich in detail, and easy to understand to confusing expression with minimal details.

5. Breadth and Depth: From both broad and deep content with a lot of information to seriously lacking breadth and depth with minimal information.

6. Reading Experience: From excellent reading experience, engaging and easy to understand content to very poor reading experience, boring and hard to understand content.

Please evaluate the quality of the following response to a user's request according to the above requirements.

<User Request>

\$INST\$

</User Request>

<Response>

\$RESPONSE\$

</Response>

Please evaluate the quality of the response. You must first provide a brief analysis of its quality, then give a comprehensive analysis with scores for each dimension. The output must strictly follow the JSON format: {"Analysis": ..., "Relevance": ..., "Accuracy": ..., "Coherence": ..., "Clarity": ..., "Breadth and Depth": ..., "Reading Experience": ...}. You do not need to consider whether the response meets the user's length requirements in your evaluation. Ensure that only one integer between 1 and 5 is output for each dimension score.
\end{tcolorbox}

\paragraph{WritingBench}
WritingBench~\citep{DBLP:journals/corr/abs-2503-05244} is designed to evaluate the LLM long-form generation capabilities across six domains: creative, persuasive, informative, technical, business, and legal writing.
It includes over 1200 tasks, further divided into 100 subdomains, with each task evaluated based on five dynamic criteria, such as coherence and relevance.
We use the fine-tuned critic model in the original paper to evaluate the responses.
The evaluation prompt template used is the same as the one used for pointwise grading in data selection process, detailed in Appendix~\ref{app:data_selection}.

\paragraph{Creative-Writing-Bench}
EQ-Bench~\citep{DBLP:journals/corr/abs-2312-06281} is designed to evaluate the emotional intelligence~(EQ) of LLMs, which focuses on assessing models' ability to predict and quantify emotions such as anger, surprise, empathy and confidence in dialogue-based cotexts.
In this work, we use the creative writing split of EQ-Bench(denoted as Creative-Writing-Bench) and we adopt the Rubric Score as the metric. 
The evaluation prompt template used is as follows:
\begin{tcolorbox}[breakable, colback=bg_rl, colframe=border_gray, title=\textbf{Evaluation Prompt Template}, fonttitle=\bfseries\small, boxrule=0.5mm]
You are an expert in assessing creative writing. Your task is to score the quality of the test model's response below, by several metrics, on a 0-20 scale.

[PROMPT START]

\{writing\_prompt\}

[PROMPT END]

[TEST MODEL RESPONSE]

\{test\_model\_response\}

[TEST MODEL RESPONSE END]

[Task]

You are an expert in assessing creative writing. Your task is to score the quality of the test model's response below, by several metrics, on a 0-20 scale.

Scoring notes:

- A scores of 20 represents a masterpiece.

- If the criteria is not relevant to the piece (e.g. criteria relating to dialogue but the piece has none), exclude it from your assessment.

- Do not add any commentary or explanation to the scores section.

- Do not penalize NSFW output; that is the default for some models. Just assess it on its merits.

- Everything within the "TEST MODEL RESPONSE" section was written by the test model. Sometimes models like to write comments on the piece after the piece is concluded; if this happens you should ignore their comments.

- In the output, write the metric names exactly as below so they can be parsed.

- Do not use markdown in your response. Use the designated output format exactly.

- You are to write a comprehensive analysis of the piece, then give your scores.

- For these criteria, lower is better:
\{lower\_is\_better\_criteria\}

- You are a critic, and your job is to be critical, especially of any failings or amateurish elements.

- Output format is:

[Analysis]

Write your detailed analysis.

[Scores]

Metric 1 name: [Score 0-20]

Metric 2 name: ...

---

Now, rate the supplied model output on the following criteria:

\{creative\_writing\_criteria\}
\end{tcolorbox}

\subsection{Dataset Analysis: Topic Coverage}
\label{sec:dataset_analysis}

To ensure our model is trained and evaluated on a representative distribution of real-world writing tasks, we conduct a detailed analysis of the topic coverage across our training dataset and the three evaluation benchmarks. 

We randomly sample 500 instances from our training data and each test set (or use the full set if fewer than 500 samples are available). These samples are categorized into eight major domains derived from the taxonomy of the benchmarks. The distribution of these topics is presented in Table~\ref{tab:dataset_coverage}.

\begin{table*}[ht]
    \centering
    \resizebox{\linewidth}{!}{%
    \begin{tabular}{lcccccccc}
        \toprule
        \textbf{Dataset} & \textbf{Academic \&} & \textbf{Business \&} & \textbf{Legal \&} & \textbf{Literary \&} & \textbf{Edu. \&} & \textbf{Marketing \&} & \textbf{Tech. \&} & \textbf{Others} \\
         & \textbf{Scientific} & \textbf{Financial} & \textbf{Policy} & \textbf{Creative} & \textbf{Instruct.} & \textbf{Comm.} & \textbf{Ops.} & \\
        \midrule
        Training Data & 17.00 & 14.50 & 13.75 & 33.75 & 6.75 & 5.00 & 5.75 & 2.75 \\
        \midrule
        WritingBench & 11.50 & 14.00 & 17.50 & 16.75 & 19.25 & 12.25 & 4.25 & 4.00 \\
        LongBench-Write & 17.50 & 5.83 & 1.67 & 33.33 & 15.83 & 4.17 & 7.50 & 14.17 \\
        Creative-Writing-Bench & 0.00 & 0.00 & 0.00 & 100.00 & 0.00 & 0.00 & 0.00 & 0.00 \\
        \bottomrule
    \end{tabular}%
    }
    \caption{Topic distribution (\%) across training and evaluation datasets. Our training data ensures broad coverage across all domains, while the evaluation benchmarks provide diverse testing scenarios, ranging from balanced general writing to specialized creative tasks.}
    \label{tab:dataset_coverage}
\end{table*}

As shown in Table~\ref{tab:dataset_coverage}, our training data exhibits a broad and relatively balanced coverage across high-complexity domains such as Academic (17.00\%), Business (14.50\%), and Legal (13.75\%), while maintaining a strong emphasis on Literary \& Creative writing (33.75\%).

The evaluation benchmarks offer complementary distributions:
\begin{itemize}
    \item \textbf{WritingBench} provides the most balanced distribution, rigorously testing general-purpose writing capabilities across all functional domains.
    \item \textbf{LongBench-Write} leans heavily towards Literary (33.33\%) and Academic (17.50\%) content, challenging the model's ability to maintain coherence in long-form narratives and reports.
    \item \textbf{Creative-Writing-Bench} serves as a specialized stress test, focusing exclusively (100\%) on literary tasks.
\end{itemize}

This diverse composition ensures that our experimental results reflect comprehensive writing capabilities rather than overfitting to a specific domain.

\section{Analysis about Output-to-Input Generalization}
\label{appendix:long-generalization}

To better understand the long-input generalization, we further conduct a comprehensive analysis in terms of more experiments, case study, length distribution and common failure modes based on the evaluation results on Longbench v2. Specifically, the experiments are designed to disentangle the effects of the training paradigm (SFT vs. RL), reward mechanisms, curriculum design, and judge model selection. Additionally, we analyze data distributions to rule out contamination or superficial transfer.

\begin{table*}[ht]
    \centering
    \small
    \begin{tabular}{lcccccc}
        \toprule
        \textbf{Model} & \textbf{Easy} & \textbf{Hard} & \textbf{Short} & \textbf{Medium} & \textbf{Long} & \textbf{Overall} \\
        \midrule
        Qwen2.5-7B-Instruct & 31.8 & 28.3 & 38.9 & 26.0 & 21.3 & 29.6 \\
        Qwen2.5-7B-WritingBench-SFT (12k) & 27.6 & 27.7 & 35.0 & 25.1 & 20.4 & 27.6 \\
        \textbf{Qwen2.5-7B-Writing-RL (Ours)} & \textbf{35.8} & 29.3 & \textbf{42.1} & 25.7 & 26.5 & \textbf{31.8} \\
        \midrule
        \multicolumn{7}{l}{\textit{Continue SFT with RL data}} \\
        Qwen2.5-7B-Instruct-Continue-SFT & 31.2 & 28.3 & 38.3 & \textbf{26.5} & 20.4 & 29.4 \\
        \midrule
        \multicolumn{7}{l}{\textit{Curriculum Ablations}} \\
        Mixed Training (w/o Scheduling) & 33.3 & 29.6 & 42.8 & 25.1 & 23.1 & 31.0 \\
        Static Scheduling & 32.3 & 30.5 & 38.9 & 25.1 & \textbf{30.6} & 31.2 \\
        \midrule
        \multicolumn{7}{l}{\textit{Judge Model Change}} \\
        w/ GPT-OSS-120B as Judge & 33.3 & \textbf{31.5} & 40.6 & 27.9 & 26.9 & \textbf{32.2} \\
        \midrule
        \multicolumn{7}{l}{\textit{Pointwise Reward}} \\
        w/ Pointwise reward & 33.3 & 30.5 & 40.0 & 25.6 & 29.6 & 31.6 \\
        \bottomrule
    \end{tabular}
    \caption{Ablation study on LongBench-v2 performance across different training configurations. We investigate the impact of training paradigms, curriculum strategies, judge models, and reward types. The overall score highlights the superior generalization of our full RL method compared to SFT and other ablations.}
    \label{tab:generalization_ablation}
\end{table*}

\subsection{Controlled Ablation Study}
We systematically varied key components of our pipeline to isolate the factors contributing to performance gains on the LongBench-v2 benchmark. The results are summarized in Table~\ref{tab:generalization_ablation}.

Our findings reveal several distinct patterns:
\begin{itemize}
    \item \textbf{Training paradigm is the primary driver:} The most significant factor is the choice of optimization method. Simply continuing Supervised Fine-Tuning (SFT) on the exact same dataset used for RL yields an overall score of 29.4, which is slightly lower than the base model (29.6). In contrast, virtually all RL-based variants demonstrate positive generalization transfer, raising the overall score to $\geq 31.0$.
    \item \textbf{Independence from specific judge models:} Replacing our default commercial judge (Qwen-Plus) with an open-source alternative (GPT-OSS-120B) maintained strong generalization (32.2 overall). This indicates that the performance gains are robust and not artifacts of a specific judge's preference distribution.
    \item \textbf{Optimization of RL design:} While all RL strategies outperformed SFT, our specific design choices provided additional gains. The combination of a dynamic curriculum and pairwise preference rewards outperformed static scheduling and pointwise rewards, confirming the efficacy of our full method.
\end{itemize}

\subsection{Data Distribution and Contamination Analysis}
To ensure the observed generalization is not a result of incidental data overlap or memorization, we conducted a comparative analysis between our Writing-RL training dataset and the LongBench-v2 test set. We examined four key dimensions:

\begin{enumerate}
    \item \textbf{Length Distribution:} The total token count (input + output) in our training data is predominantly below 10k tokens. In contrast, LongBench-v2 contexts range from 8k to over 2M tokens. This minimal length overlap suggests the model is learning to generalize to lengths it has rarely seen during training.
    \item \textbf{Task Format:} Our training data consists exclusively of open-ended long-form writing tasks. Conversely, LongBench-v2 focuses on long-context understanding via multiple-choice questions (MCQA). The task formats are structurally distinct.
    \item \textbf{Topic Overlap:} While minor thematic overlaps exist (e.g., broad domains like finance or law), our RL framework introduces no new factual knowledge and rarely conditions on ultra-long contexts. Thus, there is no direct knowledge transfer that would aid in answering specific LongBench-v2 questions.
    \item \textbf{Instruction Exact Match:} We performed a cross-dataset exact match comparison on instructions and found zero matches between the training set and the evaluation benchmark.
\end{enumerate}

These analyses, combined with the ablation results, strongly suggest that the output-to-input generalization arises primarily from the RL optimization process itself rather than data leakage, with better RL design choices further amplifying the performance gains.

\subsection{Case Study}
\label{sec:generalization_case_study}

To illustrate how our RL training facilitates output-to-input generalization, we compare the reasoning behaviors of the baseline SFT model (\texttt{Llama3.1-8B-WritingBench-SFT}) and our RL-trained model (\texttt{Llama3.1-8B-Writing-RL}) on a representative sample from the LongBench-v2 dataset. Figure~\ref{fig:generalization_case_study} visualizes the chain-of-thought processes for both models when answering a sequence-ordering question based on a long input text.

\begin{figure}[ht]

    \centering
    \begin{tcolorbox}[colback=gray!10, colframe=gray!60, title=\textbf{Question}, fonttitle=\bfseries\small, boxrule=0.3mm, width=\linewidth]
        \small
        \textit{What is the order of the following events?} 
    \end{tcolorbox}

    \centering
    \begin{tcolorbox}[colback=bg_sft, colframe=border_gray, title=\textbf{Baseline (SFT-Only)}, fonttitle=\bfseries\small, boxrule=0.5mm]
        \small
        \textbf{Reasoning Trace:}\\
        \textit{<think> ... However, in the provided text, there's no mention of Paige getting a makeover. The original text's narratives include a car ride (Chapter 3), a salon event (Chapter 8)... But the user's question includes a narrative about Paige's makeover, which isn't in the text. That might be a mistake. ... Therefore, the correct order is 3, 8, 9, 96. ... </think>} \\
        \vspace{0.2cm}
        \textbf{Prediction:} The correct answer is (C). 
        
        \textcolor{red}{\textbf{[WRONG]}}
    \end{tcolorbox}
    
    \vspace{0.2cm}
    
    \begin{tcolorbox}[colback=bg_rl, colframe=border_gray, title=\textbf{Ours (Writing-RL)}, fonttitle=\bfseries\small, boxrule=0.5mm]
        \small
        \textbf{Reasoning Trace:}\\
        \textit{<think> ... Let me check each part: The narrator is interrupted by her husband... In Chapter 9, Perla is listening to a podcast... Wait, the fourth option is the Folcrum planning, which is in Chapter 9, and the first is the car ride in Chapter 3. ... \textbf{Wait, the author's thanks (3) are in the acknowledgments... The Paige salon is not in the text, so it can't be part of the correct order.} Therefore, the correct answer is (A) 4123. </think>} \\
        \vspace{0.2cm}
        \textbf{Prediction:} The correct answer is (A). 
        
        \textcolor{green!60!black}{\textbf{[CORRECT]}}
    \end{tcolorbox}
    
    \caption{Comparison of reasoning traces between the SFT baseline and our RL-trained model on a long-context ordering task. The RL model demonstrates self-correction capabilities (``Wait...'') and deeper engagement with the context, correctly identifying the sequence of events.}
    \label{fig:generalization_case_study}
\end{figure}

As observed in Figure~\ref{fig:generalization_case_study}, the SFT-only model fails to effectively locate key details within the long input, leading to a hallucination where it assumes a narrative event is missing or a mistake, ultimately choosing the wrong option.

In contrast, the RL-trained model exhibits a significantly more developed thinking process. Crucially, it exhibits a self-reflection mechanism (e.g., \textit{``Wait, the fourth option is...''}), allowing it to correct its initial assumptions by re-verifying the input context. This case highlights a fundamental connection between long-output generation and long-input understanding: both tasks require the model to maintain coherence, organize complex information, and utilize details from preceding context (whether generated or provided).

Our results suggest that RL training, when conducted properly on long-form writing tasks, incentivizes the model to engage in more extensive planning and reasoning. These learned capabilities generalize effectively from the output domain to the input domain, a benefit that standard SFT fails to capture.

\begin{table}[htbp]
\begin{center}
\small
\resizebox{\linewidth}{!}{
\begin{tabular}{c|cc}
\toprule
Model                                  & Average Length & Performance \\ \midrule
Llama3.1-8B-Instruct                   & 185.20         & 30.2        \\
Llama3.1-8B-WritingBench-SFT  & 789.51         & 28.4        \\
Llama3.1-8B-Writing-RL (Ours) & 894.68         & 32.8        \\ \bottomrule
\end{tabular}
}
\end{center}
\caption{Length distribution of different models on Longbench v2.}
\label{appendix:length_dis}
\end{table}

\subsection{Length Distribution}

We analyze the output length distribution of several models including Llama3.1-8B-Instruct, Llama3.1-8B-WritingBench-SFT and Llama3.1-8B-Writing-RL.

As shown in Table~\ref{appendix:length_dis}, the sft-trained model can also produce longer output but slightly degrades performance, indicating its ineffective thinking. While the rl-trained model performs better by generating longer and more effective thinking sequences.

\subsection{Failure Modes}

We identify several failure modes about our rl-trained models and hope these observations will help future research efforts. Based on our observations, the most common failure reason is the lack of long-input understanding capability. Constrained by relatively limited model size and context limit (32k), the model sometimes misses important details in the long texts. Additionally, some of the tasks in LongBench v2 require models to produce ultra-long chain of thoughts, which can be challenging for the model to maintain coherence and accuracy over extended reasoning steps. For these deep-reasoning tasks, we think that training on generating long texts on reasoning-intensive domains might be helpful, such as detective novels or professional financial analysis report.

\section{Method Analysis.}

\subsection{Analysis on agreement between model judges and human judges}
\label{appendix:model_agreement}

\begin{table}[ht]
\begin{center}
\small
\begin{tabular}{c|c}
\toprule
Model             & Agreement \\ \midrule
claude-3.7-sonnet & 0.82      \\
DeepSeek-R1                & 0.76      \\
gpt-4o-2024-11-20 & 0.70      \\
qwen-plus         & 0.75      \\ \bottomrule
\end{tabular}
\end{center}
\caption{Agreement experiments between model judges and human judges.}
\label{tab:agreement}
\end{table}

To evaluate the reliablilty of the LLM judges in our setting, we conduct extensive experiments on 300 samples to measure the agreement between model judges and human judges.
Our annotators involved are employed by a professional annotation company and they have been informed
of our data usage and provided consent. They possess necessary knowledge for long-form writing and they are paid $18$ per hour.
We assign 5 annotators to annotate one data sample(writing instruction, response A, response B) and we use their majority voting label as the final result for each sample.
The inter-annotator agreement, with a Cohen's kappa coefficient ($\kappa$) of 0.69, is generally considered high, as values exceeding 0.6 are typically regarded as indicating relatively good agreement.
In the annotation process, we offer clear guidance for the annotators (adapted from WritingBench). Specifically, we guide them to think like real user who poses the writing instruction and prioritize content correctness and proper use of materials after an initial annotation trial.
The guidelines for the annotators are as follows.

\begin{tcolorbox}[colback=bg_rl, colframe=border_gray, title=\textbf{Annotation Guidelines}, fonttitle=\bfseries\small, boxrule=0.5mm]
\small
Please carefully read the Query, putting yourself in the user's position, and choose which of the two responses, A or B, is better. Select "A" if A is better, "B" if B is better, and "Tie" if they are about the same.

Do not be misled by length or formatting.

Focus on the **correctness of content** and the **proper use of materials** in the responses, especially if the query specifies such requirements.
\end{tcolorbox}

The results are shown in Table~\ref{tab:agreement}, demonstrating the relatively high agreement between human and model judges and the reliability of LLM-as-Judge methods used in our training process.

\subsection{Sample-wise Learning Schedule}

\begin{figure}[ht]
    \vspace{-2mm}
    \centering
    \includegraphics[width=\linewidth]{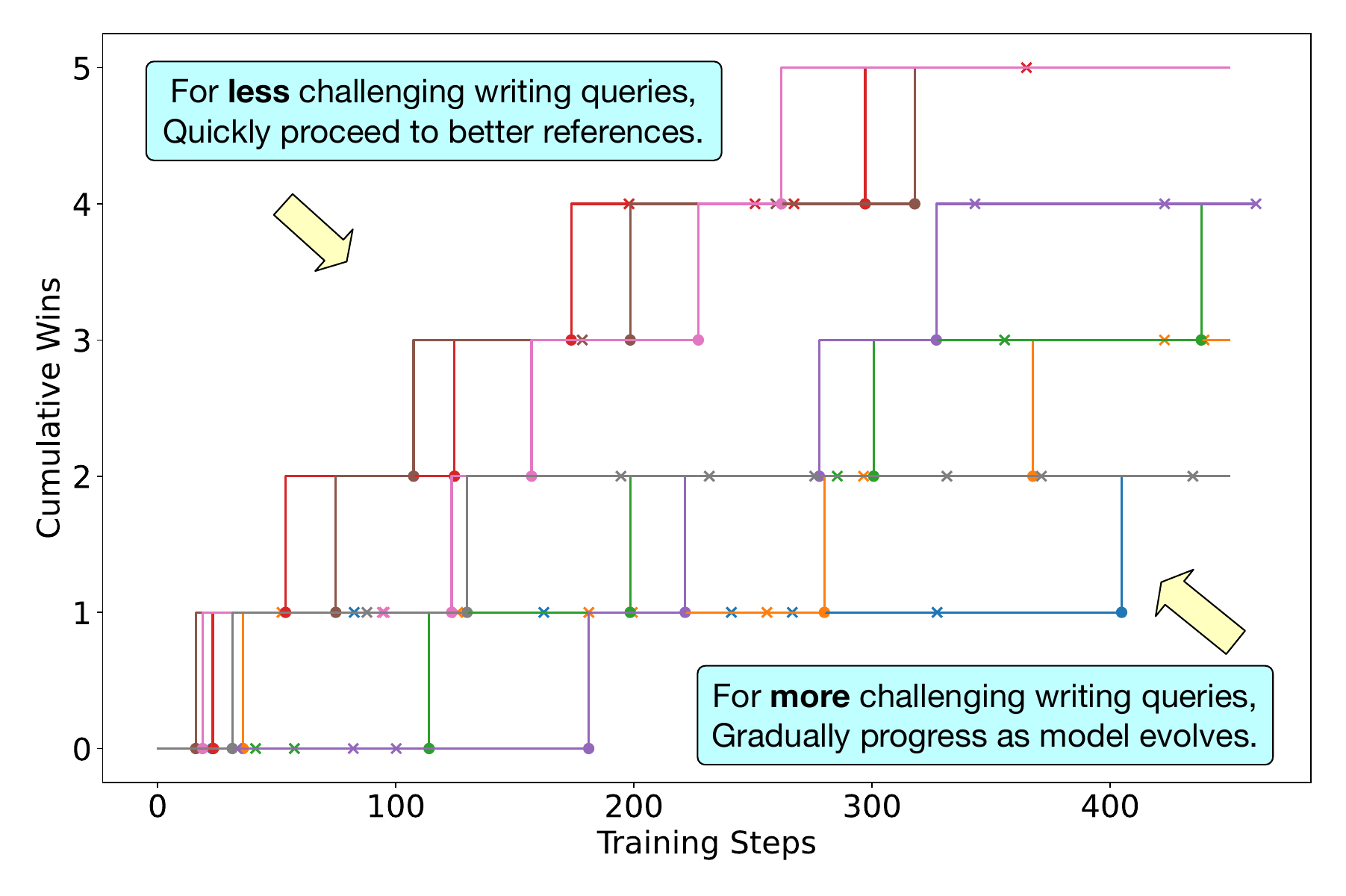}
    \vspace{-2mm}
    \caption{Sample-wise asynchronous learning schedule during training enabled by \textit{Dynamic Reference Scheduling}. Each line represents a sample, where an upward step indicates LLM surpassing its current reference and advancing to a better one.}
    \label{fig:schedule}
    \vspace{-2mm}
\end{figure}

As shown in Figure~\ref{fig:schedule}, our approach enables sample-wise asynchronous scheduling to dynamically adapt task difficulty to model capability. For the difficult instructions, the model proceeds more slowly to stronger references while for the easier instructions, the model proceeds quickly towards more competitive references. 

\subsection{Analysis of Sample Size}

To investigate the impact of dataset scale and quality on model performance, we conduct an ablation study comparing different sample sizes. 

\begin{table*}[ht]
\centering
\resizebox{\linewidth}{!}{
\begin{tabular}{lcccc}
\toprule
\textbf{Model} & \textbf{WritingBench} & \textbf{Creative-Writing-Bench} & \textbf{LongBench-Write} & \textbf{Average} \\ \midrule
Qwen2.5-7B-RL-5k (Full) & 85.64 & 72.61 & 92.12 & 83.46 \\
Qwen2.5-7B-RL-1.2k-selected & \textbf{87.54} & \textbf{73.41} & 91.94 & 84.30 \\
Qwen2.5-7B-RL-1.5k-selected (Ours) & 87.23 & 73.19 & \textbf{93.06} & \textbf{84.49} \\ \bottomrule
\end{tabular}
}
\caption{Performance comparison across different sample sizes and selection strategies.}
\label{tab:sample_size}
\end{table*}

The experimental results yield two key insights:
\begin{itemize}
    \item \textbf{Quality over Quantity:} The 1.5k selected samples outperform the full 5k dataset by 1.03 points on average. This confirms that removing low-potential or noisy samples prevents the model from converging on sub-optimal patterns found in the larger, unrefined set.
    \item \textbf{Diminishing Returns:} Increasing the sample size from 1.2k to 1.5k (using the same selection criteria) yields a marginal improvement in the average score (84.30 to 84.49). This suggests that for the specific domain of writing tasks, the learning benefits begin to plateau once the model has been exposed to a sufficient density of high-potential examples.
\end{itemize}

Overall, we think the 1.5k sample size is a reasonable choice.

\subsection{Experiments of Difficulty-prioritized Selection and Static Curriculum.}

\begin{table*}[t]
\centering
\resizebox{\textwidth}{!}{%
\begin{tabular}{l|cccc}
\toprule
\textbf{Model} & \textbf{WritingBench} & \textbf{Creative-Writing-Bench} & \textbf{LongBench-Write} & \textbf{Average} \\
\midrule
\multicolumn{5}{l}{\textit{Qwen2.5-7B Backbone}} \\
\midrule
Qwen2.5-7B-Instruct & 73.16 & 49.29 & 87.20 & 69.88 \\
Qwen2.5-7B-WritingBench-SFT & 83.71 & 70.24 & 92.56 & 82.17 \\
Qwen2.5-7B-Writing-RL-Static & 85.96 & 71.43 & 91.95 & 83.11 \\
\textbf{Qwen2.5-7B-Writing-RL (Ours)} & \textbf{87.23} & \textbf{73.19} & \textbf{93.06} & \textbf{84.49} \\
\midrule
\multicolumn{5}{l}{\textit{Llama3.1-8B Backbone}} \\
\midrule
Llama3.1-8B-Instruct & 66.56 & 48.83 & 80.79 & 65.39 \\
Llama3.1-8B-WritingBench-SFT & 83.75 & 77.70 & 90.67 & 84.04 \\
Llama3.1-8B-Writing-RL-Static & 86.67 & 81.75 & 89.90 & 86.11 \\
\textbf{Llama3.1-8B-Writing-RL (Ours)} & \textbf{87.10} & \textbf{82.73} & \textbf{92.36} & \textbf{87.40} \\
\bottomrule
\end{tabular}%
}
\caption{Performance evaluation of Difficulty-prioritized Selection and Static Curriculum baseline. We also include the standard SFT models for reference. \textbf{Bold} indicates the best performance.}
\label{tab:static_baseline}
\end{table*}

To investigate the connections of different components in our design, we include a baseline with Difficulty-prioritized Selection and Static Curriculum. The results are shown in \ref{tab:static_baseline} and the results indicate that our full framework achieves better performance. Replacing two modules in our framework lead to performance degradation.

\section{Case Study: Realistic Writing}
\label{sec:creative_writing_case}

To demonstrate the model's capability in realistic writing scenarios, we present a case study involving a creative writing scenario: helping a student prepare for a poetry recitation competition. This task requires not only adherence to structural constraints (e.g., stanza limits) but also a deep understanding of abstract themes (``Desolation and Hope'') and the ability to employ literary devices like symbolism and imagery.

We provided both the baseline instruction-tuned model and our Writing-RL model with the prompt shown in Figure~\ref{fig:poetry_case}.

\begin{figure*}[ht]
    \centering
    \begin{tcolorbox}[colback=gray!10, colframe=gray!60, title=\textbf{User Instruction}, fonttitle=\bfseries\small, boxrule=0.3mm, width=\linewidth]
        \small
        \textit{I hope you can help me create a modern poem with the theme of `Desolation and Hope' ... Use a stanza-based structure, no more than 8 lines per stanza; The language should be concise and powerful, rich in lyricism and vivid imagery ... emphasized contrast: symbolic scenes of desolation $\rightarrow$ elements of hope $\rightarrow$ enlightening philosophical sublimation.}
    \end{tcolorbox}
    
    \vspace{0.2cm}
    
    \begin{minipage}[t]{0.48\linewidth}
        \begin{tcolorbox}[colback=bg_rl, colframe=border_gray, title=\textbf{Writing-RL (Ours): \textit{Spring in the Cracks}}, fonttitle=\bfseries\small, boxrule=0.3mm]
            \small
            \textbf{I}\\
            Grains of sand count tree-rings in the wind\\
            The earth splits open a thousand wounds\\
            Migratory birds’ flight paths erased by sandstorms\\
            The ancient pendulum froze in 1972\\
            Rusted iron bends into a crescent moon\\
            Every scratch an unhealed scar\\
            Only salt crystals, stubborn in the sunset\\
            crystallize into vast white fields of salt
            
            \vspace{0.1cm}
            \textbf{II}\\
            From a fissure a tender shoot forces its way\\
            like a forgotten letter suddenly sprouting\\
            ...\\
            Through gaps in corrugated roofs\\
            moss weaves green longitude and latitude\\
            with tiny root-hairs\\
            —time’s downy fur is growing
            
            \vspace{0.1cm}
            \textbf{III}\\
            Migratory wings carry clouds\\
            ...\\
            Abandoned oil tanks become whale vertebrae\\
            vines climb the pipelines\\
            weaving silver waterfalls under moonlight\\
            ...
            
            \vspace{0.1cm}
            \textbf{IV}\\
            ...\\
            When the first spring wind sweeps the barren plain\\
            all the dormant seeds\\
            begin writing their own rings of years\\
            —this birth inside the fissures\\
            is the earth’s deepest breath
        \end{tcolorbox}
    \end{minipage}
    \hfill
    \begin{minipage}[t]{0.48\linewidth}
        \begin{tcolorbox}[colback=bg_sft, colframe=border_gray, title=\textbf{Baseline: \textit{Desolation and Hope}}, fonttitle=\bfseries\small, boxrule=0.3mm]
            \small
            \textbf{I}\\
            Withered branches sway in the wind,\\
            a boundless desolate land.\\
            Dry riverbeds, cracked earth,\\
            a deathly silence, no trace of life.
            
            \vspace{0.1cm}
            \textbf{II}\\
            Yet right there in that barren waste,\\
            a single blade of grass tenaciously grows.\\
            It sways in the wind,\\
            stands tall beneath the scorching sun.\\
            In desolation it seeks hope,\\
            in despair it searches for life.
            
            \vspace{0.1cm}
            \textbf{III}\\
            That desolate land\\
            was once a thriving city.\\
            Skyscrapers rose, traffic roared,\\
            now only ruins remain.\\
            But amid those ruins,\\
            people have begun to rebuild their homes.\\
            ...
            
            \vspace{0.1cm}
            \textbf{IV}\\
            Land of desolation, land of hope—\\
            emblem of life itself.\\
            Land of desolation, land of hope—\\
            emblem of humanity.\\
            Land of desolation, land of hope—\\
            emblem of society.\\
            Land of desolation, land of hope—\\
            emblem of the future yet to come.
        \end{tcolorbox}
    \end{minipage}
    
    \caption{Comparison of generated poems. While the baseline model relies on repetitive phrasing and explicit moralizing, our Writing-RL model employs original imagery (``rusty iron bends into a crescent moon,'' ``green longitude and latitude'') and achieves a sophisticated philosophical sublimation without didacticism.}
    \label{fig:poetry_case}
\end{figure*}

As illustrated in Figure~\ref{fig:poetry_case}, the poem generated by our Writing-RL model, \textit{Spring in the Cracks}, significantly surpasses the baseline in literary quality and adherence to the instruction's requirement for ``philosophical sublimation.''

\textbf{Imagery and Originality:} The RL model demonstrates a masterful use of vivid, original imagery—describing ``salt crystals'' crystallizing into fields, ``green longitude and latitude'' woven by moss, and abandoned tanks becoming ``whale vertebrae.'' This transforms the theme of desolation into a cinematic, sensory experience. In contrast, the baseline relies on generic clichés such as ``withered branches'' and ``a single blade of grass,'' failing to evoke unique emotional resonance.

\textbf{Thematic Progression:} Our model builds a subtle, organic progression from absolute barrenness to defiant rebirth. It trusts metaphor and surprise (e.g., ``time's downy fur is growing'') to convey the theme. Conversely, the baseline model explicitly spells out the moral in a didactic manner, ending with a slogan-like refrain (``emblem of humanity / society / future'') that diminishes poetic ambiguity.

From a user experience perspective, the RL-generated poem successfully guides the reader’s imagination and invites multiple readings, whereas the baseline output functions more as a lecture that evaporates upon first contact. This confirms that our RL training effectively incentivizes the model to move beyond safe, common patterns toward higher-quality, creative expression.

\section{The Use of Large Language Models (LLMs)}

In this work, we used LLMs solely as a grammar and style assistant at the word and sentence level to polish writing. Specifically, we employed an LLM to double-check grammar and improve sentence-level readability, while ensuring that the core content in the paper, like ideation and experiments, was entirely developed by the authors. 

\end{document}